\documentclass[twoside]{article}
\usepackage{algorithm}
\usepackage{algorithmic}
\usepackage{amssymb,amsmath,amsthm}
\usepackage{graphicx}
\usepackage{preamble}
\usepackage{natbib}
\usepackage{hyperref}
\usepackage{color}
\usepackage{wasysym}
\usepackage{subfigure}
\usepackage{tabularx}
\usepackage{booktabs}
\usepackage{bm}

\definecolor{mydarkblue}{rgb}{0,0.08,0.45}
\hypersetup{ %
    pdftitle={},
    pdfauthor={},
    pdfsubject={},
    pdfkeywords={},
    pdfborder=0 0 0,
    pdfpagemode=UseNone,
    colorlinks=true,
    linkcolor=mydarkblue,
    citecolor=mydarkblue,
    filecolor=mydarkblue,
    urlcolor=mydarkblue,
    pdfview=FitH}

\newcolumntype{x}[1]{>{\centering\arraybackslash\hspace{0pt}}m{#1}}
\newcommand{\tabbox}[1]{#1}

\definecolor{WowColor}{rgb}{.75,0,.75}
\definecolor{SubtleColor}{rgb}{0,0,.50}

\newcounter{margincounter}
\newcommand{\displaycounter}{{\arabic{margincounter}}}
\newcommand{\incdisplaycounter}{{\stepcounter{margincounter}\arabic{margincounter}}}

\newcommand{\fTBD}[1]{\textcolor{SubtleColor}{$\,^{(\incdisplaycounter)}$}\marginpar{\tiny\textcolor{SubtleColor}{ {\tiny $(\displaycounter)$} #1}}}

\newif\ifarXiv
\ifarXiv
	\usepackage[arxiv]{format/icml2013v2}
\else
	\usepackage[accepted]{format/icml2013v2}
\fi
\begin{document}

\twocolumn[
\icmltitle{Structure Discovery in Nonparametric Regression through Compositional Kernel Search}

\icmlauthor{David Duvenaud$^{*\dagger}$}{dkd23@cam.ac.uk}
%\icmladdress{University of Cambridge}
\icmlauthor{James Robert Lloyd$^{*\dagger}$}{jrl44@cam.ac.uk}
%\icmladdress{University of Cambridge}
\icmlauthor{Roger Grosse$^{\ddagger}$}{rgrosse@mit.edu}
%\icmladdress{Massachussets Institute of Technology}
\icmlauthor{Joshua B. Tenenbaum$^{\ddagger}$}{jbt@mit.edu}
%\icmladdress{Massachussets Institute of Technology}
\icmlauthor{Zoubin Ghahramani$^{\dagger}$}{zoubin@eng.cam.ac.uk}
%\icmladdress{University of Cambridge}
%\icmladdress{Brain and Cognitive Sciences, Massachusetts Institute of Technology}    
            
\icmlkeywords{nonparametrics, gaussian process, machine learning, ICML, structure learning, extrapolation, regression, kernel learning, equation learning, supervised learning, time series}
\vskip 0.3in
]

\begin{abstract}
%\fTBD{How do we get author addresses and joint authorship showing properly in ICML accepted format?}
Despite its importance, choosing the structural form of the kernel in nonparametric regression remains a black art.
We define a space of kernel structures which are built compositionally by adding and multiplying a small number of base kernels.
We present a method for searching over this space of structures which mirrors the scientific discovery process.
The learned structures can often decompose functions into interpretable components and enable long-range extrapolation on time-series datasets.
Our structure search method outperforms many widely used kernels and kernel combination methods on a variety of prediction tasks.
%Nonparametric regression methods are used widely and successfully, but their effectiveness depends strongly on choosing an appropriate measure of similarity between data points.
%This is often specified by a kernel, and while there are effective data-driven techniques for learning kernel parameters, choosing the structural form of the kernel remains a black art.
%\fTBD{Mention \gp{}s? Or generic kernel statement?}
%\NA{, often expressed as a kernel function}.
%The composite kernels in our search space are highly interpretable and can express abstract properties of a function.
%Specifically, composite kernels allow for the automatic decompositon of a function into diverse and interpetable components, and in some cases allow for long-range extrapolation.
\end{abstract}

%Nonparametric regression methods are used widely and successfully but their effectiveness depends strongly on choosing an appropriate measure of similarity between data points.
%In particular, Gaussian processes (\gp{}) require the specification of a kernel and data driven techniques to select an appropriate kernel are underdeveloped.
%As a solution we introduce a marginal-likelihood based search over composite kernel structures.
%The composite kernels in our search space are highly interpretable and can express abstract properties of a regression function.
%Specifically, the kernels allow for the automatic decompositon of a regression function into qualitatively heterogenous and interpetable components, and in some cases they allow for long-range extrapolation.
%Furthermore, our search space is sufficently rich that we achieve state-of-the-art predictive performance on interpolation tasks.

\section{Introduction}

%Supervised learning problems, such as classification and regression, learn a function $\function$ from some input (predictor) variables, $\InputVar$, to some output (response) variables, $\outputVar$.
Kernel-based nonparametric models, such as support vector machines and Gaussian processes (\gp{}s), have been one of the dominant paradigms for supervised machine learning over the last 20 years.
These methods depend on defining a kernel function, $\kernel(\inputVar,\inputVar')$, which specifies how similar or correlated outputs $\outputVar$ and $\outputVar'$ are expected to be at two inputs $\inputVar$ and $\inputVar'$.
By defining the measure of similarity between inputs, the kernel determines the pattern of inductive generalization.
%The kernel function gives a measure of similarity between inputs and determines the pattern of inductive generalizations that are made. 

Most existing techniques pose kernel learning as a (possibly high-dimensional) parameter estimation problem.
%The ability to learn kernels automatically has helped greatly in making kernel methods accessible to non-experts.
Examples include learning hyperparameters \cite{rasmussen38gaussian}, linear combinations of fixed kernels \cite{Bach_HKL}, and mappings from the input space to an embedding space \cite{salakhutdinov2008using}.

However, to apply existing kernel learning algorithms, the user must specify the parametric form of the kernel, and this can require considerable expertise, as well as trial and error.

To make kernel learning more generally applicable, we reframe the kernel learning problem as one of structure discovery, and automate the choice of kernel form.
In particular, we formulate a space of kernel structures defined compositionally in terms of sums and products of a small number of base kernel structures.
This provides an expressive modeling language which concisely captures many widely used techniques for constructing kernels.
We focus on Gaussian process regression, where the kernel specifies a covariance function, because the Bayesian framework is a convenient way to formalize structure discovery.
Borrowing discrete search techniques which have proved successful in equation discovery \cite{todorovski1997declarative} and unsupervised learning \cite{grosse2012exploiting}, we automatically search over this space of kernel structures using marginal likelihood as the search criterion.

We found that our structure discovery algorithm is able to automatically recover known structures from synthetic data as well as plausible structures for a variety of real-world datasets. 
On a variety of time series datasets, the learned kernels yield decompositions of the unknown function into interpretable components that enable accurate extrapolation beyond the range of the observations.
Furthermore, the automatically discovered kernels outperform a variety of widely used kernel classes and kernel combination methods on supervised prediction tasks.

%\fTBD{I hate these paragraphs, but they seem to be required...} 
%Section \ref{sec:Structure} outlines some commonly used kernel families as well as ways in which they can be composed. 
%Our grammar over kernels and our proposed structure discovery algorithm are described in Section \ref{sec:Search}. 
%Section \ref{sec:related_work} situates our work in the context of other nonparametric regression, kernel learning, and structure discovery methods.
%We evaluate our methods on synthetic datasets, time series analysis, and high-dimensional prediction problems in Sections \ref{sec:synthetic} through \ref{sec:quantitative}, respectively.

While we focus on Gaussian process regression, we believe our kernel search method can be extended to other supervised learning frameworks such as classification or ordinal regression, or to other kinds of kernel architectures such as kernel SVMs.
We hope that the algorithm developed in this paper will help replace the current and often opaque art of kernel engineering with a more transparent science of automated kernel construction%\fTBD{Previously discovery}
.

\section{Expressing structure through kernels} 
\label{sec:Structure}

%Kernel functions $\kernel : \InputSpace \times \InputSpace \to \Reals$ can be used to define a measure of similarity between two points $\inputVar, \inputVar'$ in some space $\InputSpace$.
%\fTBD{Define $\GP\sim\dots$ here?}
Gaussian process models use a kernel to define the covariance between any two function values: ${\textrm{Cov}(\outputVar, \outputVar') = \kernel(\inputVar,\inputVar')}$.
%The kernel determines which sorts of structures the model places most of its probability mass upon, and in effect determines the generalization properties of the model.
The kernel specifies which structures are likely under the \gp{} prior, which in turn determines the generalization properties of the model.
%The kernel, $\kernel$, must define a valid covariance function\fTBD{expand me}; when this is the case $\kernel$ is said to be positive semi-definite (PSD).
%\fTBD{RBG: we should make the distinction between kernels and kernel families, and then use it consistently}
%
%Examples of commonly used kernels include squared exponential (SE), periodic (Per) and linear kernels (Lin) defined below\fTBD{If we add RQ then we have everything in one place}
%\begin{eqnarray}
%\kernel_\textrm{SE}(\inputVar, \inputVar') = & \sigma^2\exp\left(-\frac{(\inputVar - \inputVar')^2}{2\ell^2}\right) \\
%\kernel_\textrm{Per}(\inputVar, \inputVar') = & \sigma^2\exp\left(-\frac{2\sin^2(\pi|\inputVar - \inputVar'|/p)}{\ell^2}\right) \\
%\kernel_\textrm{Lin}(\inputVar, \inputVar') = & \sigma_b^2 + \sigma_v^2(\inputVar - \ell)(\inputVar' - \ell).
%\end{eqnarray}
%
In this section, we review the ways in which kernel families\footnotemark can be composed to express diverse priors over functions. 
\footnotetext{When unclear from context, we use `kernel family' to refer to the parametric forms of the functions given in the appendix. A kernel is a kernel family with all of the parameters specified.}  

There has been significant work on constructing \gp{} kernels and analyzing their properties, summarized in Chapter 4 of \cite{rasmussen38gaussian}. 
Commonly used kernels families include the squared exponential (\kSE), periodic (\kPer), linear (\kLin), and rational quadratic (\kRQ) (see Figure~\ref{fig:basic_kernels} and the appendix).
\newcommand{\fhbig}{1.6cm}
\newcommand{\fwbig}{1.8cm}
\newcommand{\kernpic}[1]{\includegraphics[height=\fhbig,width=\fwbig]{figures/structure_examples/#1}}
\newcommand{\kernpicr}[1]{\rotatebox{90}{\includegraphics[height=\fwbig,width=\fhbig]{figures/structure_examples/#1}}}
\newcommand{\addkernpic}[1]{{\includegraphics[height=\fhbig,width=\fwbig]{figures/additive_multi_d/#1}}}
\newcommand{\largeplus}{\tabbox{{\Large+}}}
\newcommand{\largeeq}{\tabbox{{\Large=}}}
\newcommand{\largetimes}{\tabbox{{\Large$\times$}}}
\begin{figure}[ht]
\centering
\renewcommand{\tabularxcolumn}[1]{>{\arraybackslash}m{#1}}
%\begin{tabular}{m{\fwbig}m{0.01\textwidth}m{\fwbig}m{0.01\textwidth}m{\fwbig}m{\fwbig}m{\fwbig}}
%\begin{tabular}{C{\fwbig}C{\fwbig}C{\fwbig}C{\fwbig}}%{m{\fwbig}m{\fwbig}m{\fwbig}}
\begin{tabularx}{\columnwidth}{XXXX}
%Composite & Draws from \gp{} & \gp{} posterior \\ \toprule
  \kernpic{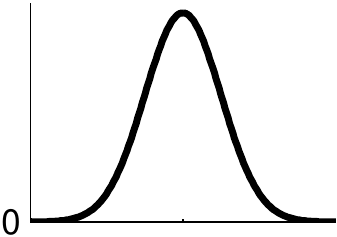} & \kernpic{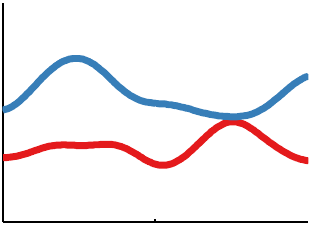}
& \kernpic{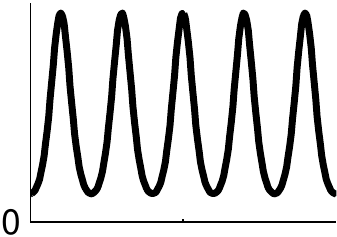} & \kernpic{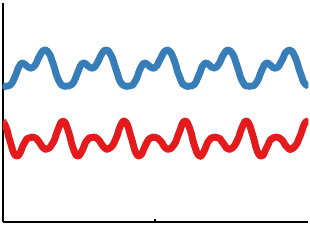}
\\
  {\small Squared-exp (\kSE)} & {\small local \newline variation} 
& {\small Periodic (\kPer)} & {\small repeating structure}
\\
\midrule
  \kernpic{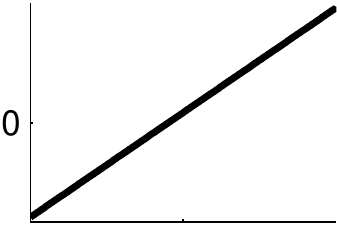} & \kernpic{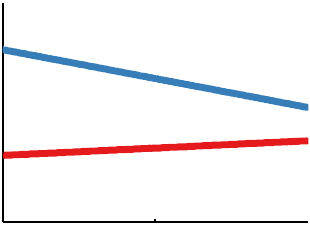}
& \kernpic{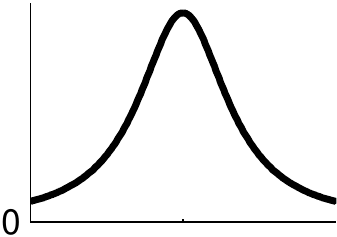} & \kernpic{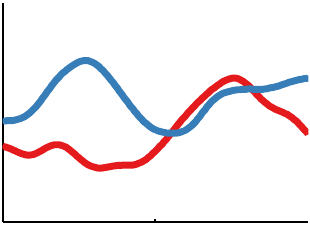}
\\
  {\small Linear (\kLin)} & {\small linear \newline functions} 
& {\small Rational- \newline quadratic(\kRQ)} & {\small multi-scale \newline \phantom{iii}variation}
\end{tabularx}
\caption{ 
%Examples of structures expressible by composite kernels.  
%The x-axis has the same scale for all plots.
  Left and third columns: base kernels $k(\cdot,0)$.  Second and fourth columns: draws from a \gp{} with each repective kernel.  The x-axis has the same range on all plots.}
\label{fig:basic_kernels}
\end{figure}

%\vspace{-0.5cm}
\paragraph{Composing Kernels}
Positive semidefinite kernels (\ie those which define valid covariance functions) are closed under addition and multiplication.
% \ie any algebraic composition of PSD kernels will define a PSD kernel.\fTBD{Cite theorem}
This allows one to create richly structured and interpretable kernels from well understood base components.
%Figure \ref{fig:kernels} shows several examples of structured kernels that can be constructed by adding or multiplying standard base kernels.

All of the base kernels we use are one-dimensional; kernels over multidimensional inputs are constructed by adding and multiplying kernels over individual dimensions.
These dimensions are represented using subscripts, e.g. $\SE_2$ represents an \kSE{} kernel over the second dimension of $\inputVar$.
\begin{figure}[ht]
\centering
\renewcommand{\tabularxcolumn}[1]{>{\arraybackslash}m{#1}}
%\begin{tabular}{m{\fwbig}m{0.01\textwidth}m{\fwbig}m{0.01\textwidth}m{\fwbig}m{\fwbig}m{\fwbig}}
%\begin{tabular}{C{\fwbig}C{\fwbig}C{\fwbig}C{\fwbig}}%{m{\fwbig}m{\fwbig}m{\fwbig}}
\begin{tabularx}{\columnwidth}{XXXX}
  \kernpic{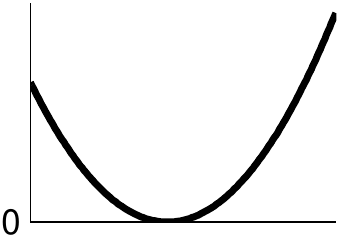} & \kernpic{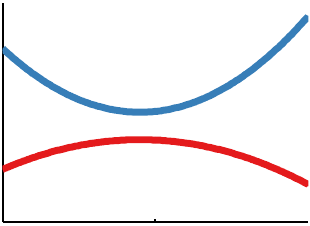} 
& \kernpic{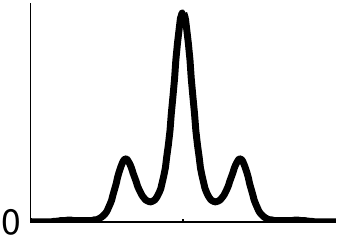} & \kernpic{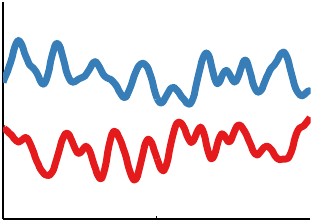}
\\
  {\small $\kLin \times \kLin$} & {\small quadratic functions}
& {\small $\kSE \times \kPer$} & {\small locally \newline periodic}
\\
\midrule 
  \kernpic{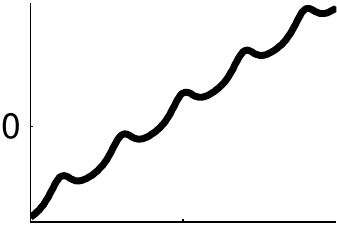} & \kernpic{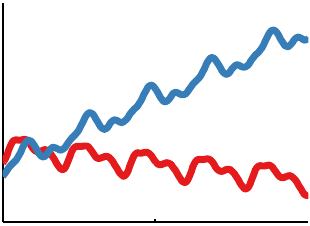}
& \kernpic{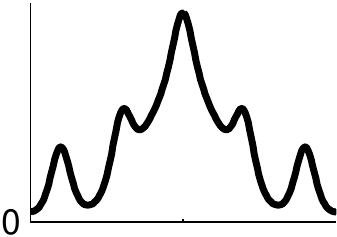} & \kernpic{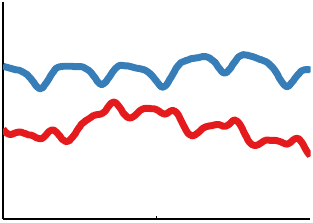}
\\
  {\small $\kLin + \kPer$} & {\small periodic with trend}
& {\small $\kSE + \kPer$ } & {\small periodic with noise}
\\
\midrule
  \kernpic{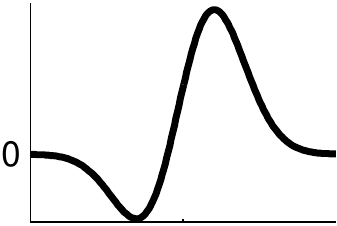} & \kernpic{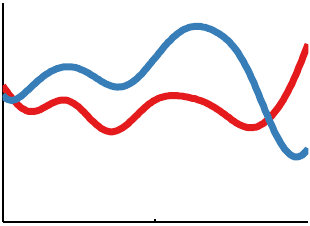}
& \kernpic{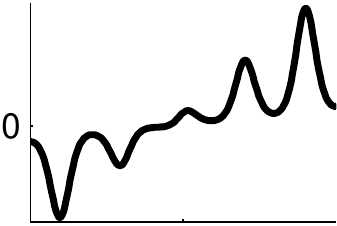} & \kernpic{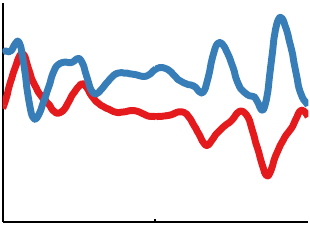}
\\
  {\small $\kLin \times \kSE$} & {\small increasing variation}
& {\small $\kLin \times \kPer$} & {\small growing amplitude}
\\
\midrule
  \addkernpic{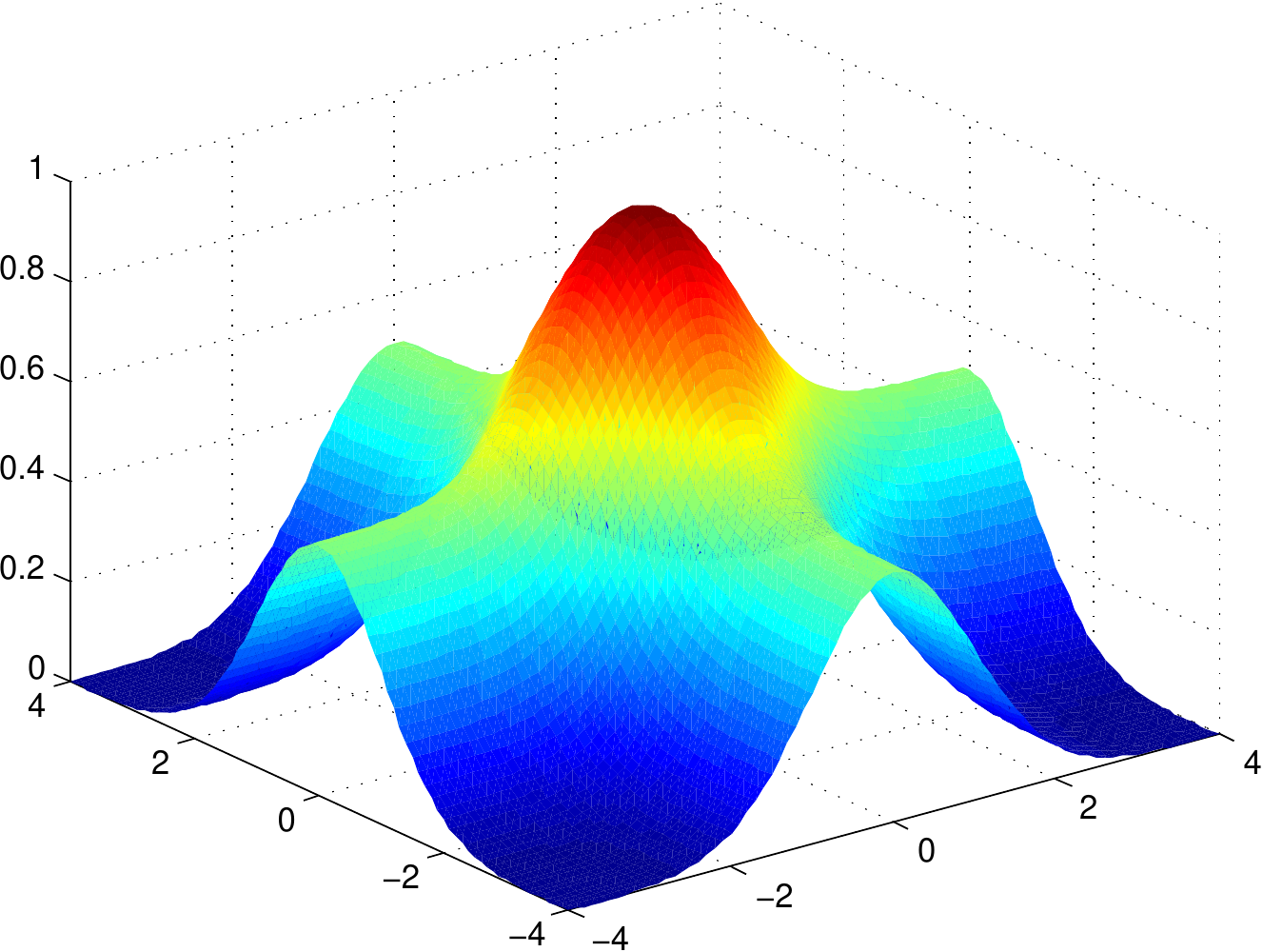} & \addkernpic{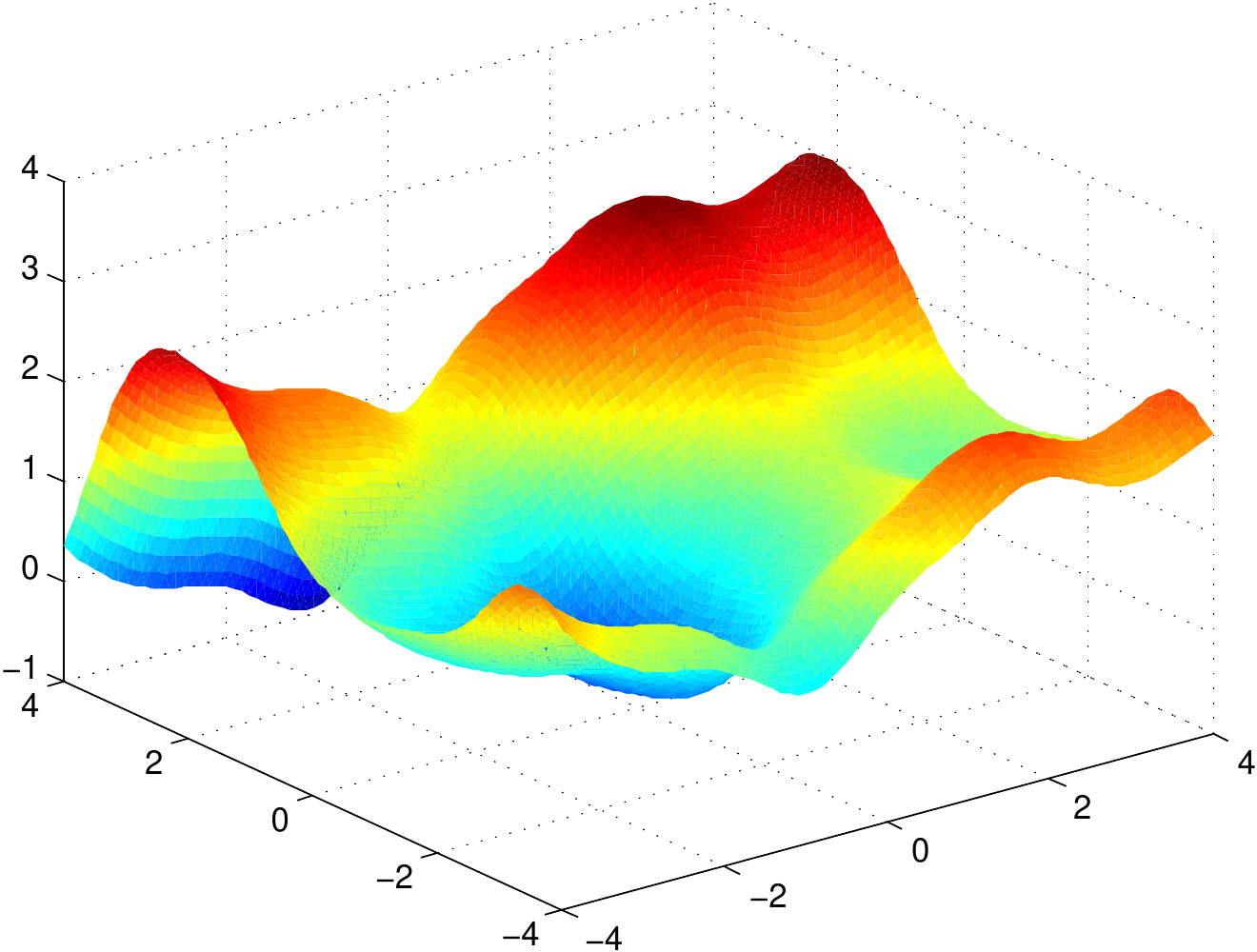}
& \addkernpic{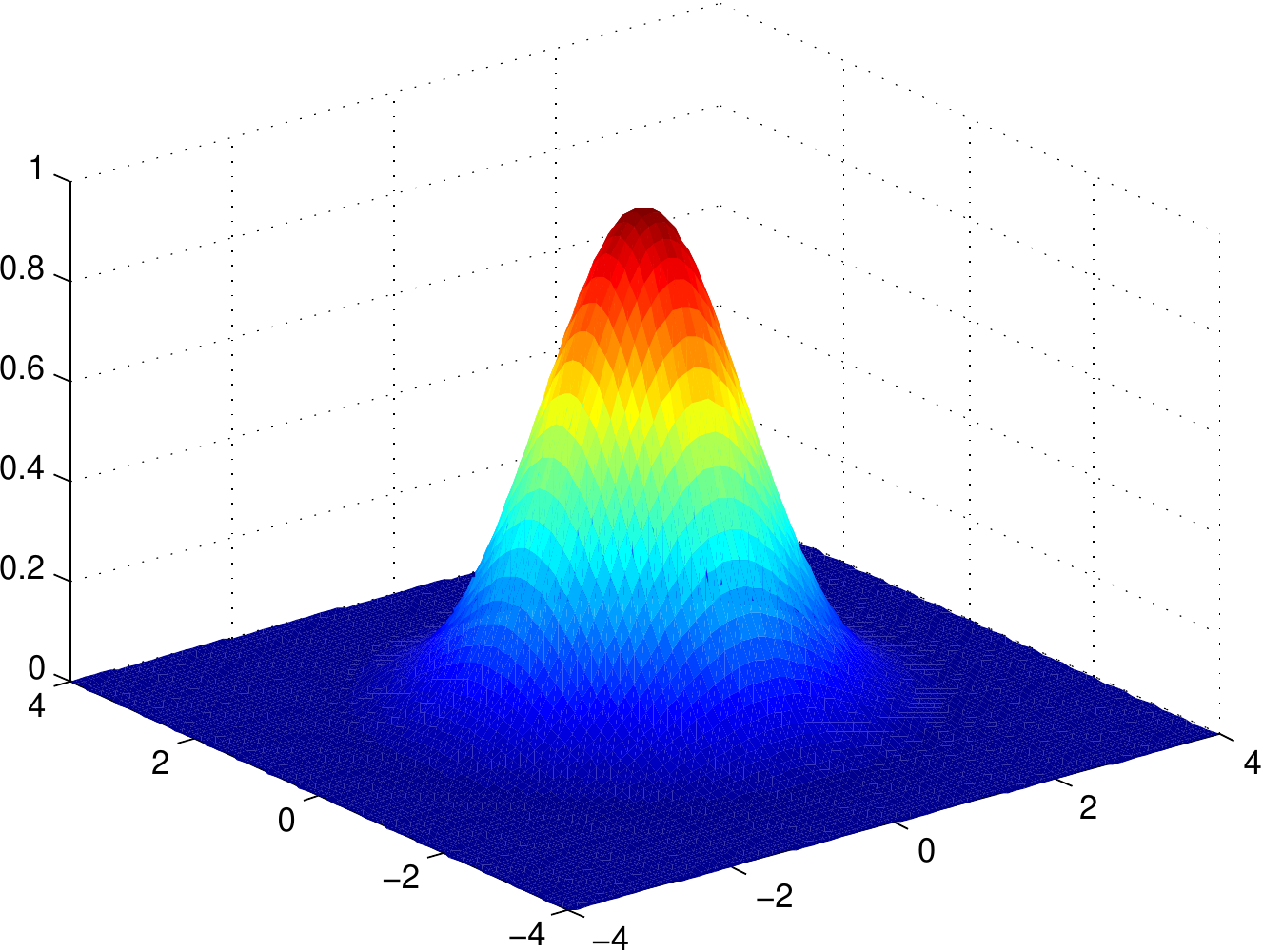}  & \addkernpic{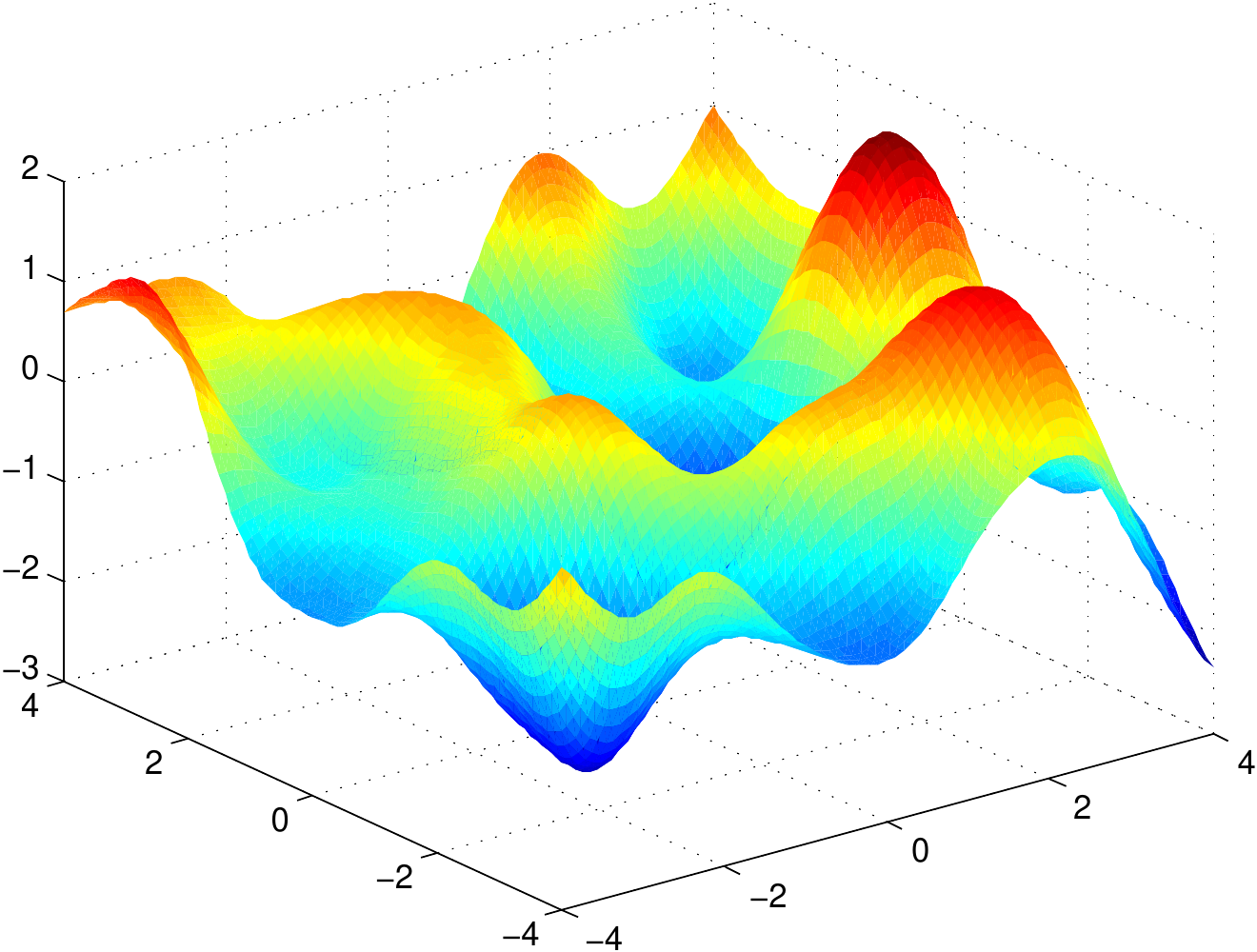}
\\
  {\small $\kSE_1 + \kSE_2$} & {\small $f_1(x_1)$ $+ f_2(x_2)$}
& {\small $\kSE_1 \times \kSE_2$} & {\small $f(x_1, x_2)$}
\end{tabularx}
\caption{ Examples of structures expressible by
  composite kernels.  
%The x-axis has the same scale for all plots.
  Left column and third columns: composite kernels $k(\cdot,0)$.  Plots have same meaning as in Figure \ref{fig:basic_kernels}.}
\label{fig:kernels}
\end{figure}

\paragraph{Summation}

By summing kernels, we can model the data as a superposition of independent functions, possibly representing different structures.
Suppose functions ${\function_1, \function_2}$ are draw from independent \gp{} priors, ${\function_1 \dist \GP(\mu_1, \kernel_1)}$, ${\function_2 \dist \GP(\mu_2, \kernel_2)}$.
Then ${\function := \function_1 + \function_2 \dist \GP(\mu_1 + \mu_2, \kernel_1 + \kernel_2)}$.

In time series models, sums of kernels can express superposition of different processes, possibly operating at different scales.
In multiple dimensions, summing kernels gives additive structure over different dimensions, similar to generalized additive models~\citep{hastie1990generalized}.
These two kinds of structure are demonstrated in rows 2 and 4 of figure~\ref{fig:kernels}, respectively.

\paragraph{Multiplication}

Multiplying kernels allows us to account for interactions between different input dimensions or different notions of similarity. 
For instance, in multidimensional data, the multiplicative kernel $\SE_1 \times \SE_3$ represents a smoothly varying function of dimensions 1 and 3 which is not constrained to be additive.
In univariate data, multiplying a kernel by \kSE{} gives a way of converting global structure to local structure. 
For example, $\Per$ corresponds to globally periodic structure, whereas $\Per \times \SE$ corresponds to locally periodic structure, as shown in row 1 of figure~\ref{fig:kernels}.

Many architectures for learning complex functions, such as convolutional networks \cite{lecun1989backpropagation} and sum-product networks \cite{poon2011sum}, include units which compute AND-like and OR-like operations.
Composite kernels can be viewed in this way too. A sum of kernels can be understood as an OR-like operation: two points are considered similar if either kernel has a high value.
Similarly, multiplying kernels is an AND-like operation, since two points are considered similar only if both kernels have high values.
Since we are applying these operations to the similarity functions rather than the regression functions themselves, compositions of even a few base kernels are able to capture complex relationships in data which do not have a simple parametric form.

\paragraph{Example expressions}

In addition to the examples given in Figure~\ref{fig:kernels}, many common motifs of supervised learning can be captured using sums and products of one-dimensional base kernels:

\begin{tabular}{l|l}
Bayesian linear regression & $\Lin$ \\
%Bayesian quadratric regression & $\Lin \times \Lin$ \\
Bayesian polynomial regression & $\Lin \times \Lin \times \ldots$\\
Generalized Fourier decomposition & $\Per + \Per + \ldots$ \\
Generalized additive models & $\sum_{d=1}^D \SE_d$ \\
Automatic relevance determination & $\prod_{d=1}^D \SE_d$ \\
Linear trend with local deviations & $\Lin + \SE$ \\
Linearly growing amplitude & $\Lin \times \SE$
\end{tabular}

We use the term `generalized Fourier decomposition' to express that the periodic functions expressible by a \gp{} with a periodic kernel are not limited to sinusoids.

%\section{Gaussian Processes Priors}

%Gaussian processes are a flexible and tractable prior over functions, useful for solving regression and classification tasks\cite{rasmussen38gaussian}.
%The kind of structure which can be captured by a GP model is mainly determined by its \emph{kernel}: the covariance function.
%One of the main difficulties in specifying a Gaussian process model is in choosing a kernel which can represent the structure present in the data.
%For small to medium-sized datasets, the kernel has a large impact on modeling efficacy.
%\TBD{Note: The above paragraph is plagarized from my additive GP paper. -David} 

%
%The technique of constructing composite kernels using sums and products of existing kernels is not new \cite{rasmussen38gaussian} [more cites, Phil Hennig's astronomy work?].  
%However, the main contribution of this paper is to automate the search over kernel structures.

\section{Searching over structures}
\label{sec:Search}

As discussed above, we can construct a wide variety of kernel structures compositionally by adding and multiplying a small number of base kernels.
In particular, we consider the four base kernel families discussed in Section \ref{sec:Structure}: \kSE, \kPer, \kLin, and \kRQ.
Any algebraic expression combining these kernels using the operations $+$ and $\times$ defines a kernel family, whose parameters are the concatenation of the parameters for the base kernel families. 

Our search procedure begins by proposing all base kernel families applied to all input dimensions. 
We allow the following search operators over our set of expressions:
\begin{itemize}
\item[(1)] Any subexpression $\subexpr$ can be replaced with $\subexpr + \baseker$, where $\baseker$ is any base kernel family.
\item[(2)] Any subexpression $\subexpr$ can be replaced with $\subexpr \times \baseker$, where $\baseker$ is any base kernel family.
\item[(3)] Any base kernel $\baseker$ may be replaced with any other base kernel family $\baseker^\prime$.
\end{itemize}

These operators can generate all possible algebraic expressions.
To see this, observe that if we restricted the $+$ and $\times$ rules only to apply to base kernel families, we would obtain a context-free grammar (CFG) which generates the set of algebraic expressions.
However, the more general versions of these rules allow more flexibility in the search procedure, which is useful because the CFG derivation may not be the most straightforward way to arrive at a kernel family.

Our algorithm searches over this space using a greedy search: at each stage, we choose the highest scoring kernel and expand it by applying all possible operators.
%\NA{
%It is unlikely that a greedy search will be optimal, but empirically it has performed well to date.
%}
%\NA{
%We then select the highest scoring kernel over the entire search.
%\footnotemark
%}
%\footnotetext{\NA{Since search operator (3) results in copies of the current composite kernel, the search typically stops itself, selecting the same kernel expression at each level of the search.}}

Our search operators are motivated by strategies researchers often use to construct kernels.
In particular,
\begin{itemize}
\item One can look for structure, \eg periodicity, in the residuals of a model, and then extend the model to capture that structure.
This corresponds to applying rule (1).
\item One can start with structure, \eg linearity, which is assumed to hold globally, but find that it only holds locally.
This corresponds to applying rule (2) to obtain the structure shown in rows 1 and 3 of figure~\ref{fig:kernels}.
\item One can add features incrementally, analogous to algorithms like boosting, backfitting, or forward selection.
This corresponds to applying rules (1) or (2) to dimensions not yet included in the model.
\end{itemize}

\paragraph{Scoring kernel families}

Choosing kernel structures requires a criterion for evaluating structures.
We choose marginal likelihood as our criterion, since it balances the fit and complexity of a model \citep{rasmussen2001occam}.  Conditioned on kernel parameters, the marginal likelihood of a \gp{} can be computed analytically.  However, to evaluate a kernel family we must integrate over kernel parameters.  We approximate this intractable integral with the Bayesian information criterion \citep{schwarz1978estimating} after first optimizing to find the maximum-likelihood kernel parameters.
%In a fully Bayesian approach, we would put priors over the parameters and compute the marginal likelihood of the models with all the parameters integrated out.
%However, as this would be difficult to do across our space of models, we approximate this integral by choosing the parameters to optimize the marginal likelihood, and then apply the Bayesian information criterion (BIC) to penalize model complexity. \fTBD{If true, say we got similar results using Laplace}
%To avoid an expensive integration over kernel parameters, we used the Bayesian information criterion \citep{schwarz1978estimating} as an approximation.
%We note that other model selection criteria could be used with our search procedure.
%For instance, random cross-validation could be used when the goal is interpolation.

Unfortunately, optimizing over parameters is not a convex optimization problem, and the space can have many local optima.
%Unfortunately, the required optimization over parameters is not convex, and the space can have many local optima.
%The example in figure \ref{fig:mauna_grow} can be modeled with a short length scale and small noise or a long length scale and large noise, and both explanations are better than any intermediate ones.
For example, in data with periodic structure, integer multiples of the true period (\ie harmonics) are often local optima. 
To alleviate this difficulty, we take advantage of our search procedure to provide reasonable initializations: all of the parameters which were part of the previous kernel are initialized to their previous values.
All parameters are then optimized using conjugate gradients, randomly restarting the newly introduced parameters.
This procedure is not guaranteed to find the global optimum, but it implements the commonly used heuristic of iteratively modeling residuals.

\section{Related Work}
\label{sec:related_work}

%\paragraph{Composite kernels in GP models} The technique of constructing composite kernels using sums and products of existing kernels was demonstrated in detail in Chapter 5 of \cite{rasmussen38gaussian}, where the resulting posterior mean was also decomposed into a sum of component-wise means, although the posterior variance was not.  While \cite{rasmussen38gaussian} manually explored several composite kernels for a particular dataset, our work automates this search process over a grammar of possible composite kernels.

%\paragraph{Gaussian process kernels}
%There has been significant work on constructing \gp{} kernels and analyzing their properties.
%This work is summarized in Chapter 4 of \cite{rasmussen38gaussian}. 

\paragraph{Nonparametric regression in high dimensions}
Nonparametric regression methods such as splines, locally weighted regression, and \gp{} regression are popular because they are capable of learning arbitrary smooth functions of the data.
Unfortunately, they suffer from the curse of dimensionality: it is very difficult for the basic versions of these methods to generalize well in more than a few dimensions.
Applying nonparametric methods in high-dimensional spaces can require imposing additional structure on the model.

One such structure is additivity.
Generalized additive models (GAM) assume the regression function is a transformed sum of functions defined on the individual dimensions: $\expect[f(\vx)] = g\inv(\sum_{d=1}^D f_d(x_d))$.
%Generalized additive models \cite{hastie1990generalized} are models in which the function is modeled as a sum of functions defined on the individual dimensions: $\expect[f(\vx)] = \sum_{d=1}^D f_d(x_d)$.
These models have a limited compositional form, but one which is interpretable and often generalizes well.
In our grammar, we can capture analogous structure through sums of base kernels along different dimensions.

It is possible to add more flexibility to additive models by considering higher-order interactions between different dimensions. 
Additive Gaussian processes \cite{duvenaud2011additive11} are a \gp{} model whose kernel implicitly sums over all possible products of one-dimensional base kernels.  
\citet{plate1999accuracy} constructs a \gp{} with a composite kernel, summing an \kSE{} kernel along each dimension, with an SE-ARD kernel (\ie a product of \kSE{} over all dimensions).
Both of these models can be expressed in our grammar.

A closely related procedure is smoothing-splines ANOVA \cite{wahba1990spline, gu2002smoothing}.
This model is a linear combinations of splines along each dimension, all pairs of dimensions, and possibly higher-order combinations.
Because the number of terms to consider grows exponentially in the order, in practice, only terms of first and second order are usually considered.

Semiparametric regression \citep[e.g.][]{ruppert2003semiparametric} attempts to combine interpretability with flexibility by building  a composite model out of an interpretable, parametric part (such as linear regression) and a `catch-all' nonparametric part (such as a \gp{} with an SE kernel).
In our approach, this can be represented as a sum of \kSE{} and \kLin{}.

\paragraph{Kernel learning}
There is a large body of work attempting to construct a rich kernel through a weighted sum of base kernels \citep[e.g.][]{christoudias2009bayesian, Bach_HKL}.
%, including \gp{} models. 
%\citet{Bach_HKL} uses a regularized optimization framework to learn a weighted sum over an exponential number of kernels.
While these approaches find the optimal solution in polynomial time, speed comes at a cost: the component kernels, as well as their hyperparameters, must be specified in advance.

Another approach to kernel learning is to learn an embedding of the data points. 
\citet{lawrence2005probabilistic} learns an embedding of the data into a low-dimensional space, and constructs a fixed kernel structure over that space.
%\NA{
This model is typically used in unsupervised tasks and requires an expensive integration or optimisation over potential embeddings when generalizing to test points.
%}
\citet{salakhutdinov2008using} use a deep neural network to learn an embedding;
%\NA{
this is a flexible approach to kernel learning but relies upon finding structure in the input density, p(\inputVar).
Instead we focus on domains where most of the interesting structure is in \function(\inputVar).
%}\fTBD{removed comment about interpretability}%; this is a flexible approach to kernel learning but potentially less interpretable.

%To learn periodic structure, 
\citet{WilAda13} derive kernels of the form ${\SE \times \cos(x-x')}$, forming a basis for stationary kernels.
% through spectral density estimation.
%motivated as spectral density estimation.
These kernels share similarities with ${\kSE \times \kPer}$ but can express negative prior correlation, and could usefully be included in our grammar.

\citet{diosan2007evolving} and \citet{bing2010gp} learn composite kernels for support vector machines and relevance vector machines, using genetic search algorithms.
%Our work goes beyond this prior work by demonstrating the structure implied by composite kernels, employing a Bayesian search criterion, and allowing for the automatic discovery of interpretable structure from data.
Our work employs a Bayesian search criterion, and goes beyond this prior work by demonstrating the interpretability of the structure implied by composite kernels, and how such structure allows for extrapolation.
%We build upon this work by demonstrating the structure implied by composite kernels, allowing for the automatic discovery of interpretable structure from data.
%We build upon this work by interpreting the structure of composite kernels, demonstrating the automatic discovery of interpretable structure from data.
%We build upon this work by 
%pointing out and demonstrating the interpretability of the structure implied by composite kernels, and demonstrating useful decompositions on several real datasets, as well as by examining extrapolation, structure recovery, and multidimensional regression.
%Our work extends theirs by examining the interpretable structure of the discovered kernels.

\paragraph{Structure discovery}

There have been several attempts to uncover the structural form of a dataset by searching over a grammar of structures. For example, \cite{schmidt2009distilling}, \cite{todorovski1997declarative} and \cite{washio1999discovering} attempt to learn parametric forms of equations to describe time series, or relations between quantities. Because we learn expressions describing the covariance structure rather than the functions themselves, we are able to capture structure which does not have a simple parametric form.

\citet{kemp2008discovery} learned the structural form of a graph used to model human similarity judgments.
Examples of graphs included planes, trees, and cylinders.
Some of their discrete graph structures have continous analogues in our own space; \eg $\SE_1 \times \SE_2$ and $\SE_1 \times \Per_2$ can be seen as mapping the data to a plane and a cylinder, respectively.

\citet{grosse2012exploiting} performed a greedy search over a compositional model class for unsupervised learning, using a grammar and a search procedure which parallel our own. This model class contained a large number of existing unsupervised models as special cases and was able to discover such structure automatically from data. Our work is tackling a similar problem, but in a supervised setting.

\section{Structure discovery in time series}
\label{sec:time_series}

%In this section, we show on several time series datasets both the kernel expression found by our search, and the complete \gp{} posterior distribution implied by that kernel and the data.  
%We then also plot that same kernel and posterior decomposed into additive components.

To investigate our method's ability to discover structure, we ran the kernel search on several time-series.
%in order to test whether it learns plausible and interpretable structure. 
%For these datasets, we show both the best kernel expressions found in each stage of the search, and the complete posterior distribution implied by that kernel and the data. 
%Additionally, we show decompositions of the time series into superpositions of components as implied by the kernel structure.
%We pay special attention to the models' extrapolations beyond the range of observed data, since this is a strong test of the correctness of the learned structure.
%\NA{
%Our method discovers rich structure in these datasets, and produces plausible extrapolations.  
%}

%We further demonstrate that the discovered structures can be decomposed into sums of interpretrable components.  
%All kernels in our search space can equivalently be written as sums of products of base kernels by applying distributivity.
%For example,
As discussed in section 2, a \gp{} whose kernel is a sum of kernels can be viewed as a sum of functions drawn from component \gp{}s.
This provides another method of visualizing the learned structures.
In particular, all kernels in our search space can be equivalently written as sums of products of base kernels by applying distributivity. For example,
\[{\SE \times (\RQ + \Lin) = \SE \times \RQ + \SE \times \Lin}.\]
We visualize the decompositions into sums of components using the formulae given in the appendix.
%By decomposing a kernel into a sum we are able to decompose .%, even when the learned kernel was not originally of this form.
%
%these decompositions can be interpreted as superpositions of causal processes.
%Details of computations are given in the appendix.
%
%\section{Decomposing a function}
%
%Sums of kernels were shown in section~\ref{sec:Structure} to correspond to sums of independent functions.
%By splitting kernel expressions found by our search algorithm into additive components, we can also decompose the \gp{} posterior over functions into a joint distribution over functions that are summed togther.
%This potentially allows us to probabilistically separate causally the independent processes that give rise to the data. 
%
%\TBD{RBG: It's not immediately obvious that this section is part of the experiments. Also, maybe consider moving it till after the quantitative comparisons?  It could be useful rhetorically to first ask, ``how well does our algorithm do,'' and then ``let's find out why it works.''}
%
The search was run to depth 10, using the base kernels from Section \ref{sec:Structure}.
%\NA{
%We consider this set of kernels since they are diverse, interpretable and through composition span a large space of priors on functions.
%}
%\footnotetext{\NA{Typically the highest scoring kernel was found at an earlier depth. We ran to a fixed depth to reduce the chance of the search stopping due to poor choices of random initial parameters for new kernels.}}

%\paragraph{Decomposing a superposition}

%In cases where the kernel expression contains any product of a sum, our software automatically expands out the expression into an equivalent sum of products.  This operation transforms the posterior into a more interpretable, but equivalent, form.

\label{sec:extrapolation}
\paragraph{Mauna Loa atmospheric CO$\mathbf{_{2}}$}

Using our method, we analyzed records of carbon dioxide levels recorded at the Mauna Loa observatory.
Since this dataset was analyzed in detail by \citet{rasmussen38gaussian}, we can compare the kernel chosen by our method to a kernel constructed by human experts.
%it serves as a test of our algorithm's ability to recover known structure.
%First, we revisit a dataset explored in \mbox{\cite{rasmussen38gaussian}}%, pages 120-126
%, where a kernel was hand-tailored to fit a \gp{} model to the dataset.

\begin{figure}[H]
\centering
\newcommand{\wmg}{0.34\columnwidth}  % width maunu growth
\newcommand{\hmg}{3.2cm}  % height maunu growth
\begin{tabular}{c}
\hspace{-0.3cm} \includegraphics[width=\wmg,height=\hmg]{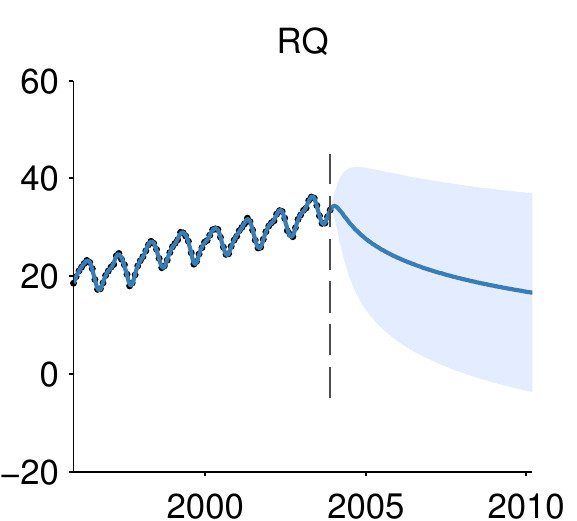} 
\hspace{-0.3cm} \includegraphics[width=\wmg,height=\hmg]{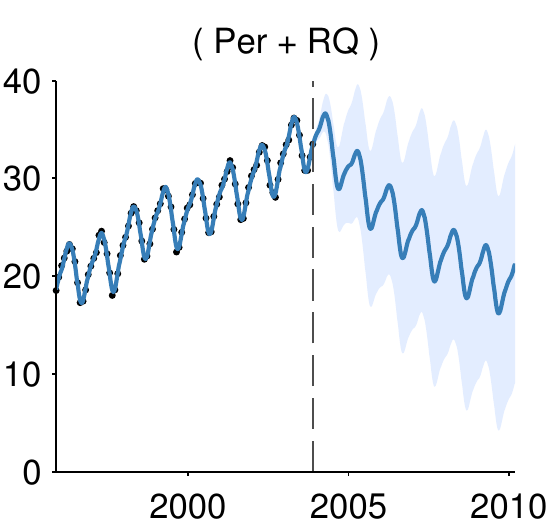}
\hspace{-0.3cm} \includegraphics[width=\wmg,height=\hmg]{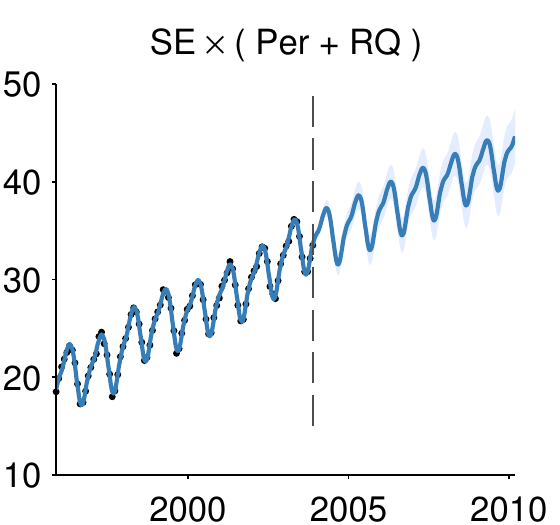}
\end{tabular}
\caption{Posterior mean and variance for different depths of kernel search.  The dashed line marks the extent of the dataset.  In the first column, the function is only modeled as a locally smooth function, and the extrapolation is poor.  Next, a periodic component is added, and the extrapolation improves.  At depth 3, the kernel can capture most of the relevant structure, and is able to extrapolate reasonably. %\TBD{RBG: (1) I think we somehow need to visualize the lengthscales, to make it obvious that the SE kernels really mean different things. (2) Why isn't SE + PE the correct answer?}
}
\label{fig:mauna_grow}
\end{figure}

\begin{figure}[H]
\newcommand{\wmgd}{1.02\columnwidth}  % width mauna decomp
\newcommand{\hmgd}{3.51cm}  % height mauna decomp
\newcommand{\mdrd}{figures/decomposition/11-Feb-03-mauna2003-s}  % mauna decomp results dir
\newcommand{\mbm}{\hspace{-0.2cm}}  % move back
\begin{tabular}{c}
\mbm \includegraphics[width=\wmgd,height=\hmgd]{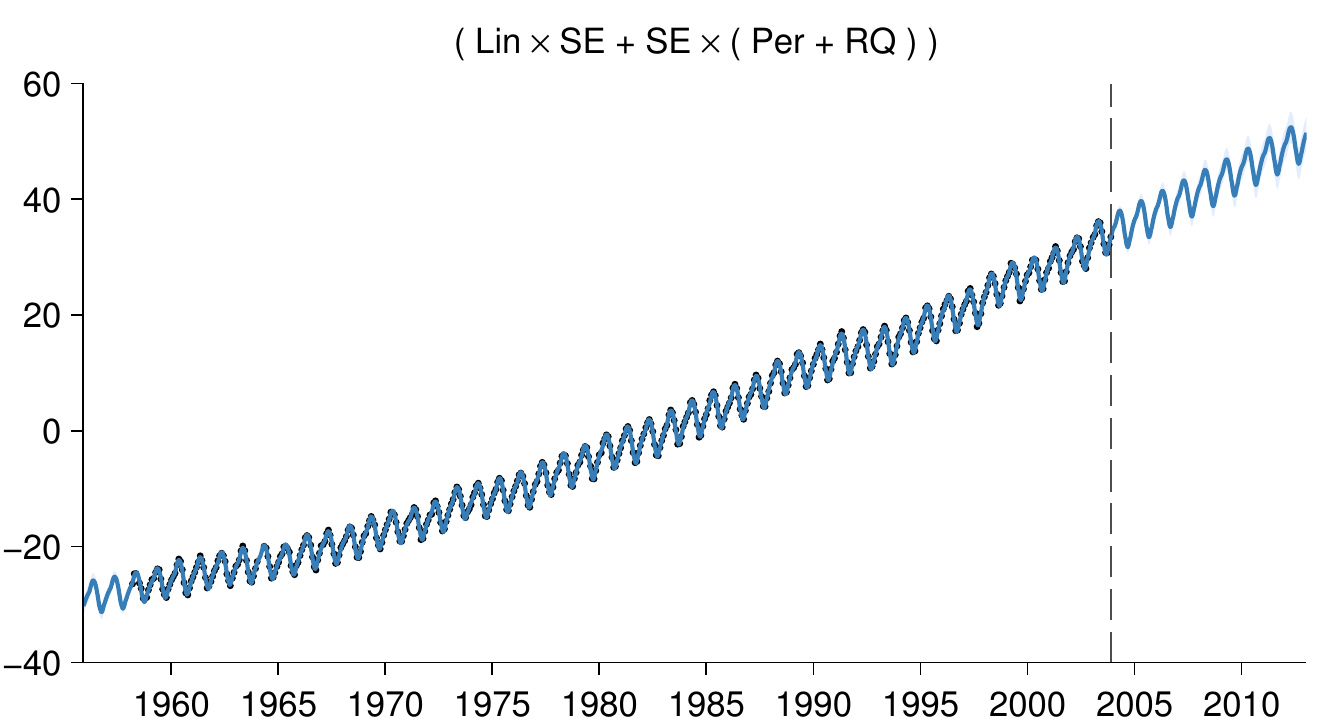} \\ = \\
\mbm \includegraphics[width=\wmgd,height=\hmgd]{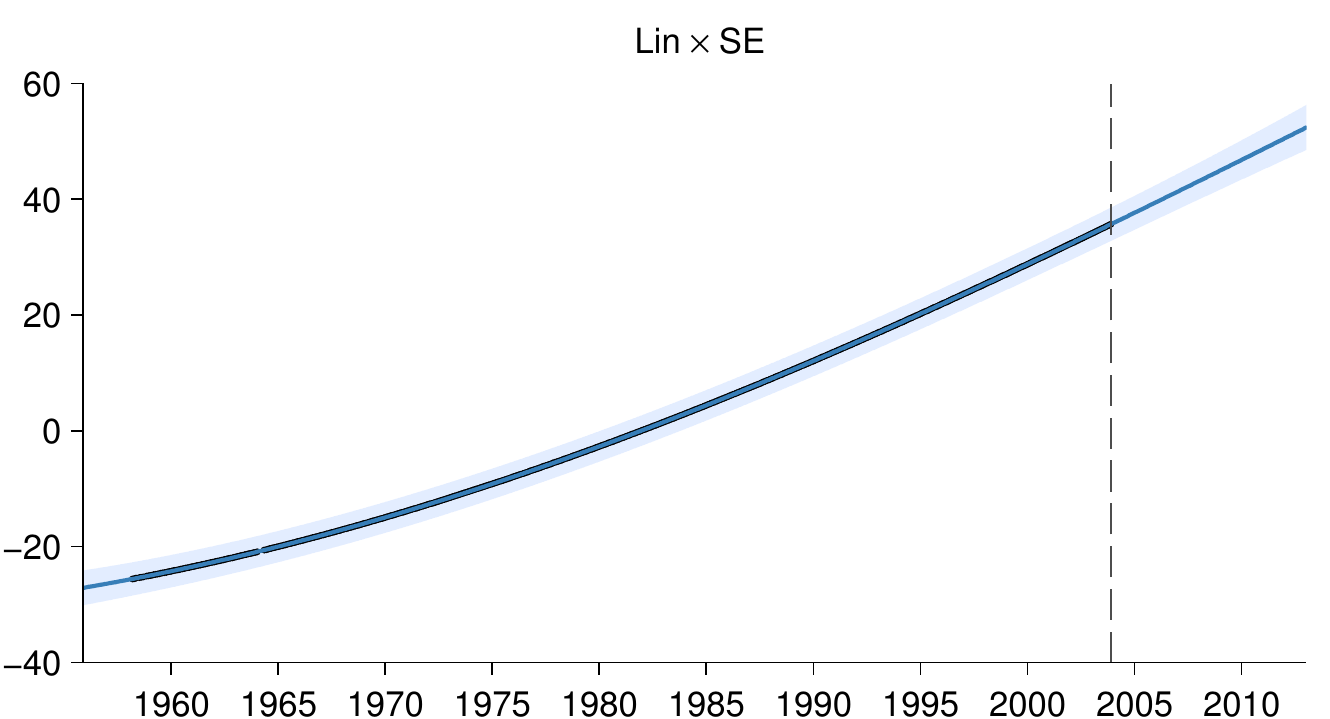} \\ + \\
\mbm \includegraphics[width=\wmgd,height=\hmgd]{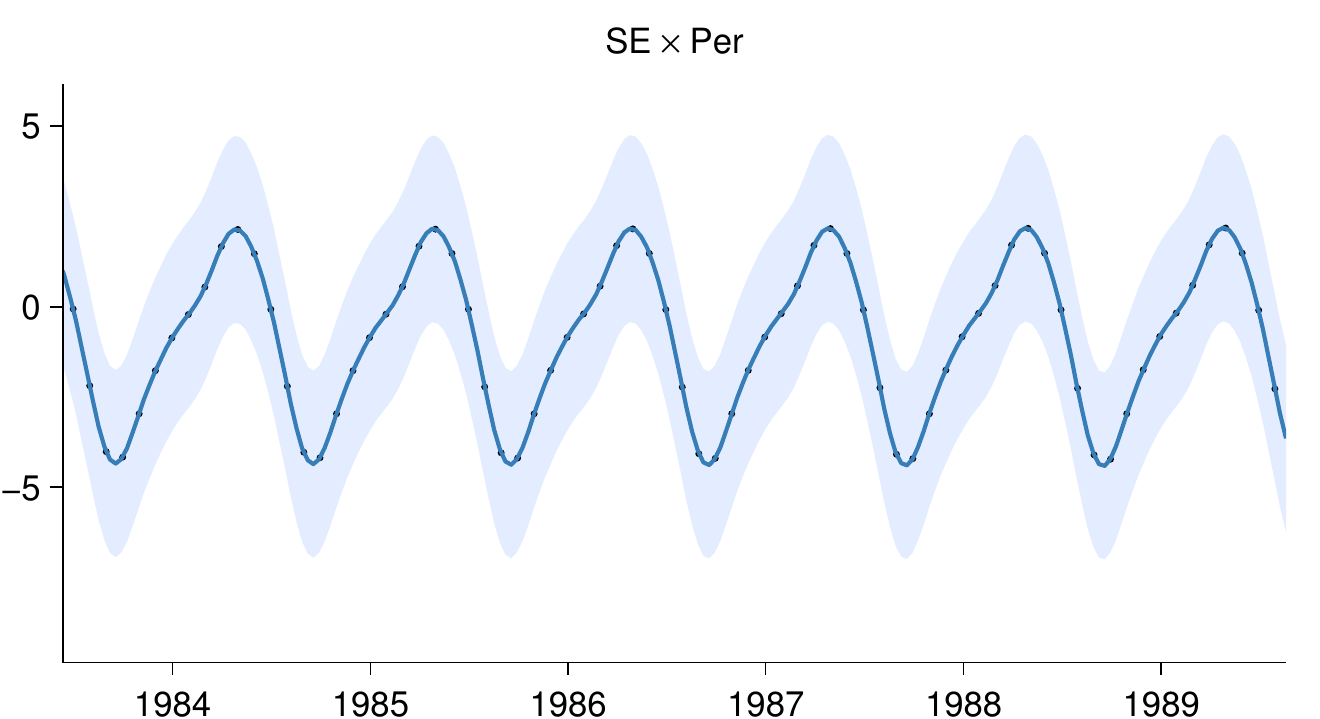} \\ + \\
\mbm \includegraphics[width=\wmgd,height=\hmgd]{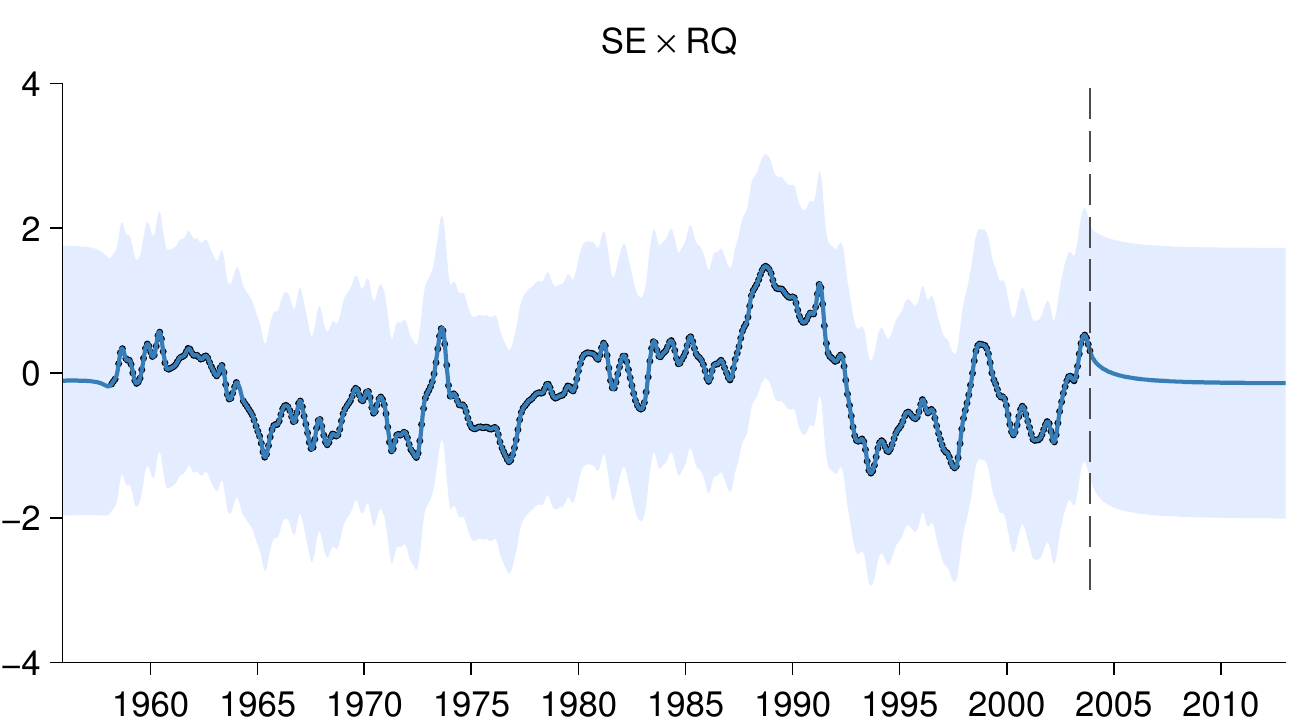} \\ + \\
\mbm \includegraphics[width=\wmgd,height=\hmgd]{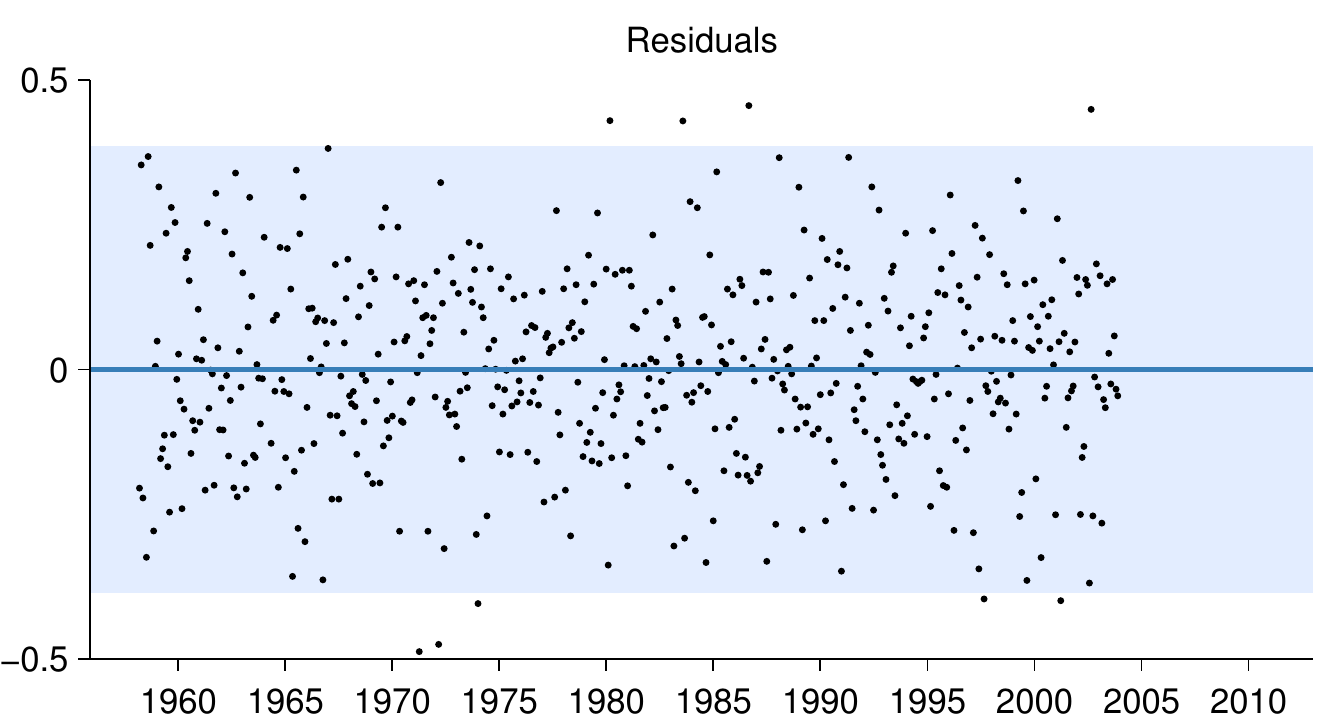}
\end{tabular}
\caption{First row: The posterior on the Mauna Loa dataset, after a search of depth 10.  Subsequent rows show the automatic decomposition of the time series.  The decompositions shows long-term, yearly periodic, medium-term anomaly components, and residuals, respectively.  In the third row, the scale has been changed in order to clearly show the yearly periodic structure.}
\label{fig:mauna_decomp}
\end{figure}

Figure \ref{fig:mauna_grow} shows the posterior mean and variance on this dataset as the search depth increases.
While the data can be smoothly interpolated by a single base kernel model, the extrapolations improve dramatically as the increased search depth allows more structure to be included.

Figure \ref{fig:mauna_decomp} shows the 
%complete posterior of the 
final model chosen by our method, together with its decomposition into additive components.
The final model exhibits both plausible extrapolation and interpretable components: a long-term trend, annual periodicity and medium-term deviations; the same components chosen by \citet{rasmussen38gaussian}.
%The final model exhibits the same structure 
%The automatically chosen kernel contains the same components as \citet{rasmussen38gaussian}: a long-term trend, annual periodicity and medium-term deviations from the trend.
We also plot the residuals, observing that there is little obvious structure left in the data.  
%
%\NA{
%
%The decomposition is qualitatively identical to that constructed in \citet{rasmussen38gaussian}.
%On this example, our search procedure is able to automate this construction where previously two \gp{} experts devoted 4 pages of text and analysis to the development of a composite kernel.
%}

\paragraph{Airline passenger data}

\begin{figure}[H]
\centering
\newcommand{\wagd}{1.02\columnwidth}  % width airline decomp
\newcommand{\hagd}{3.6cm}  % height airline decomp
\newcommand{\mb}{\hspace{-0.2cm}}  % move back
\newcommand{\ard}{figures/decomposition/31-Jan-v301-airline-months}  % airline results dir
\begin{tabular}{c}
\mb \includegraphics[width=\wagd,height=\hagd]{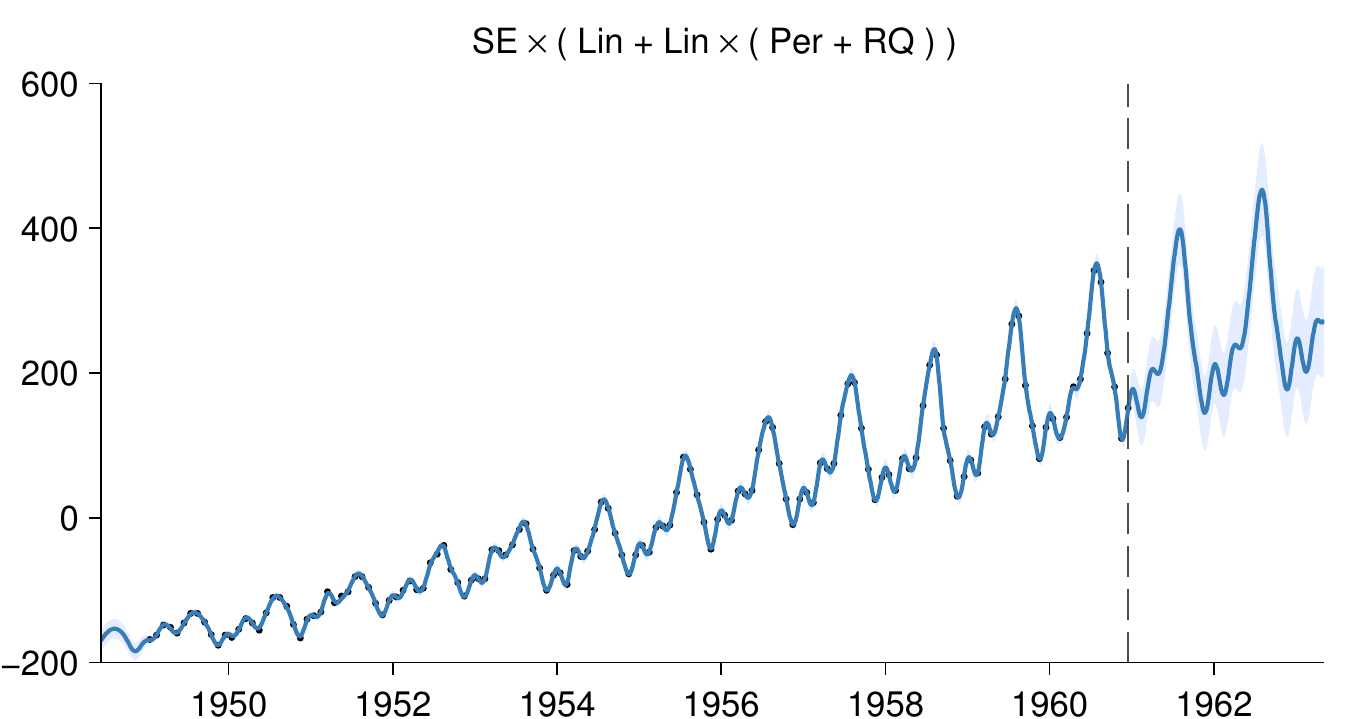} \\
 = \\ 
\mb \includegraphics[width=\wagd,height=\hagd]{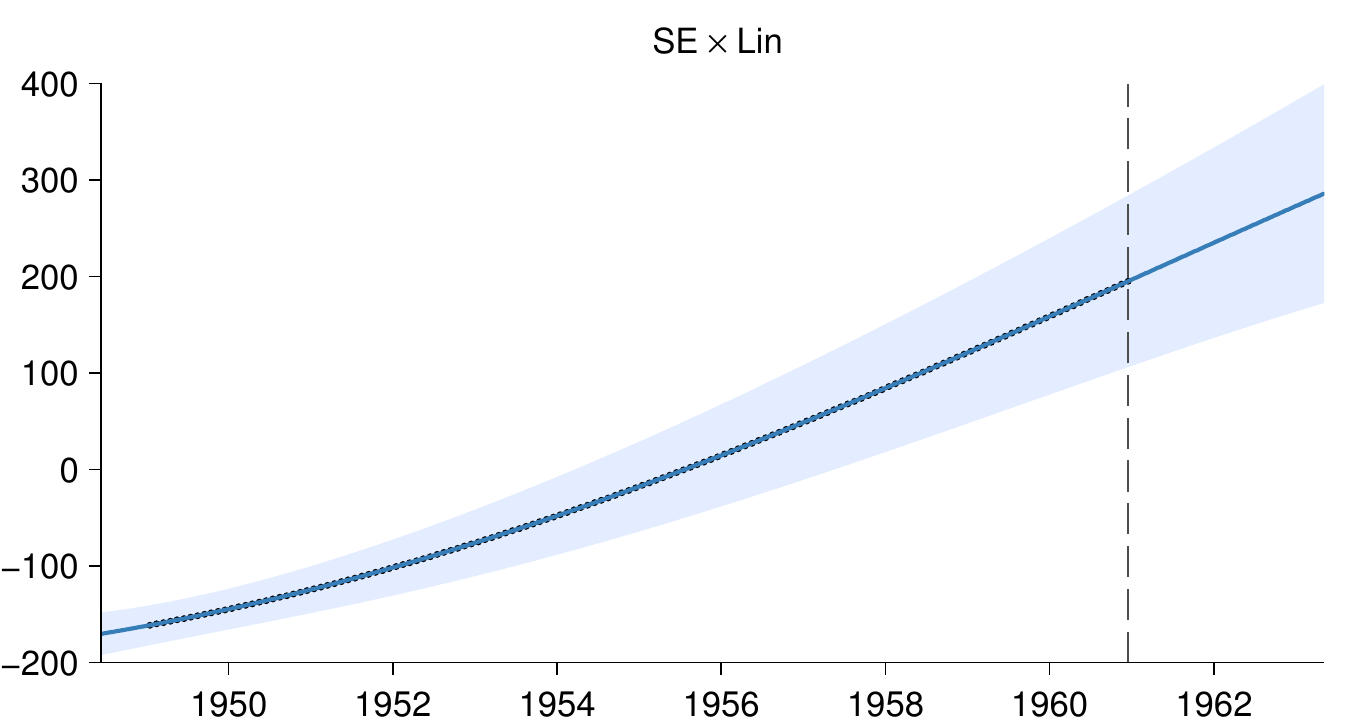} \\
 + \\
\mb \includegraphics[width=\wagd,height=\hagd]{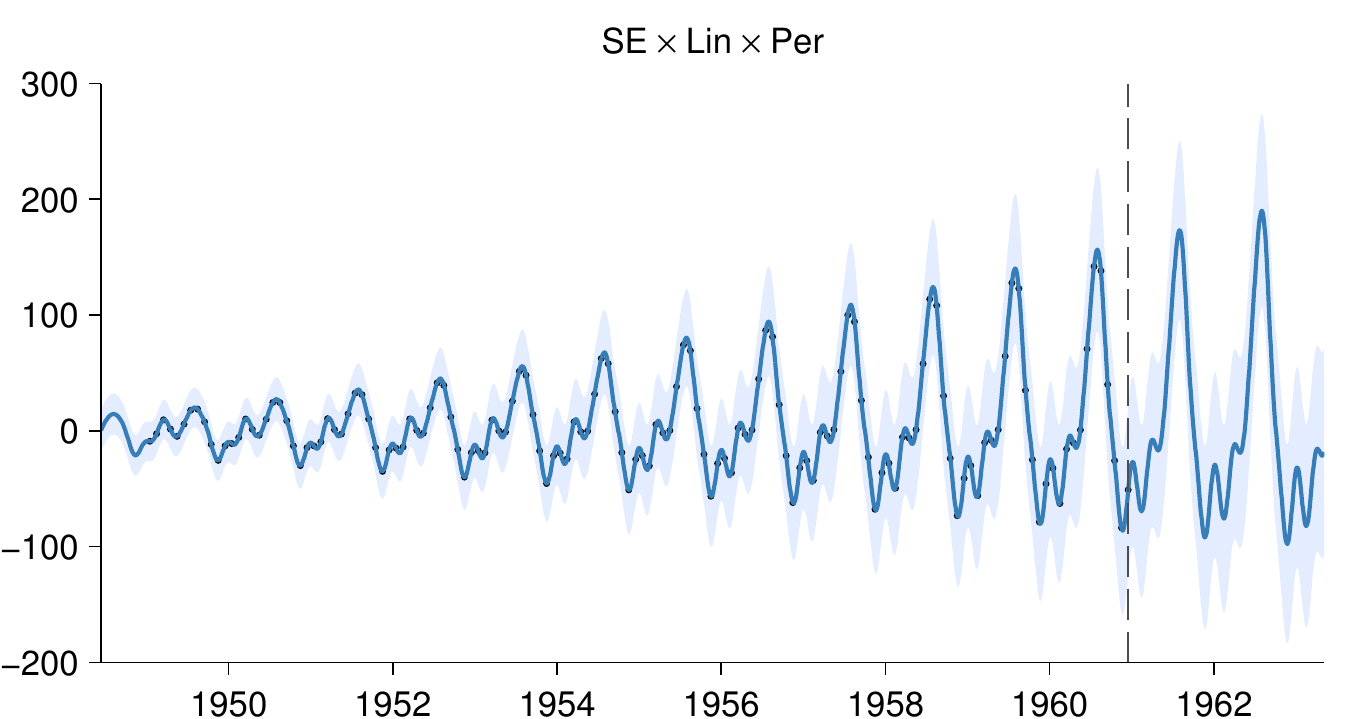} \\
 + \\
\mb \includegraphics[width=\wagd,height=\hagd]{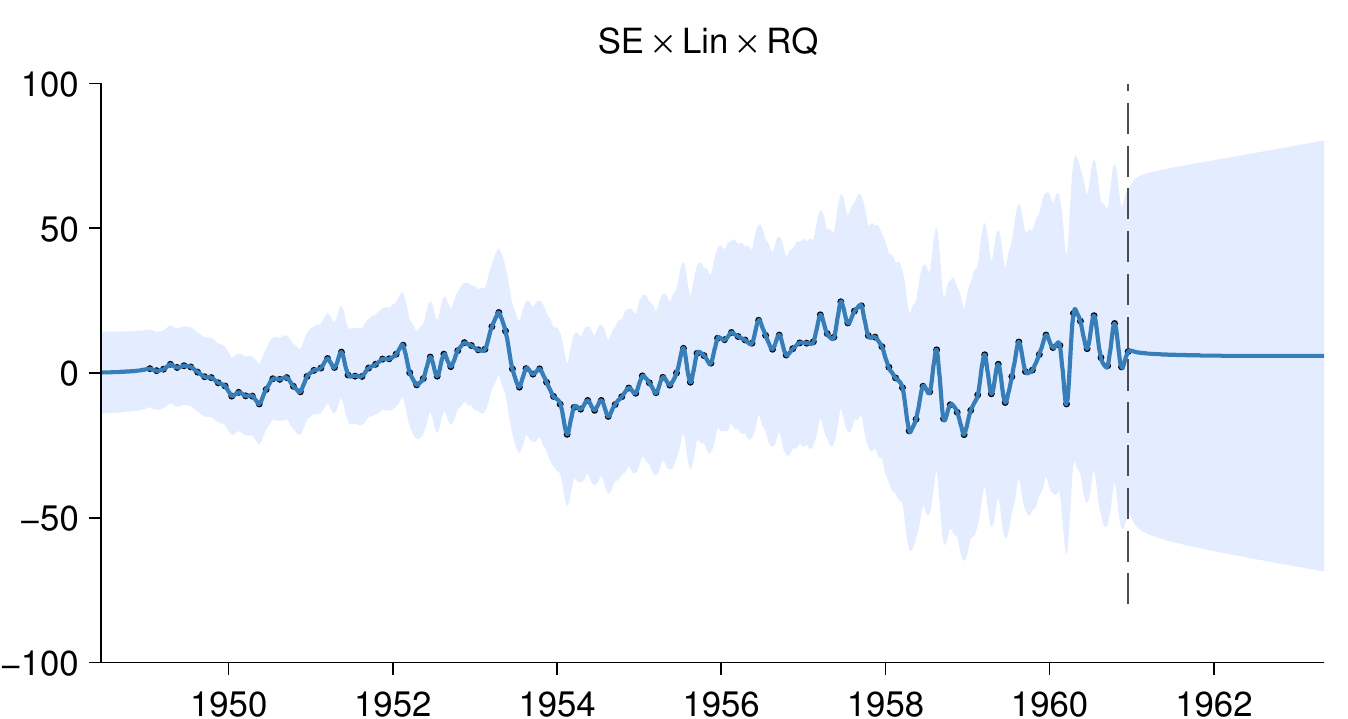} \\
 + \\
\mb \includegraphics[width=\wagd,height=\hagd]{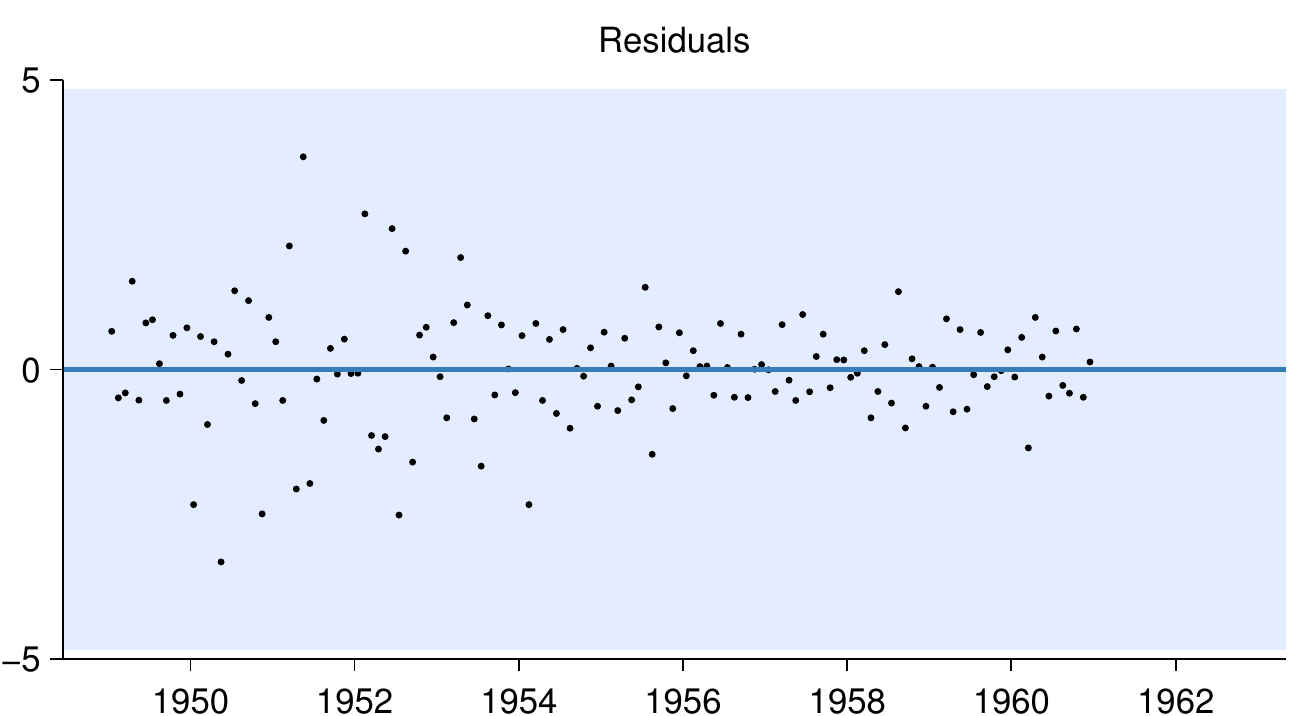}
\end{tabular}
\caption{First row:  The airline dataset and posterior after a search of depth 10.  Subsequent rows: Additive decomposition of posterior into long-term smooth trend, yearly variation, and short-term deviations.  Due to the linear kernel, the marginal variance grows over time, making this a heteroskedastic model. 
%\TBD{RBG: We allow heteroskedastic noise in our models?}
}
\label{fig:airline_decomp}
\end{figure}

Figure \ref{fig:airline_decomp} shows the decomposition produced by applying our method to monthly totals of international airline passengers~\citep{box2011time}.
We observe similar components to the previous dataset: a long term trend, annual periodicity and medium-term deviations.
In addition, the composite kernel captures the near-linearity of the long-term trend, and the linearly growing amplitude of the annual oscillations.

\paragraph{Solar irradiance Data} 
Finally, we analyzed annual solar irradiation data from 1610 to 2011 \citep{lean1995reconstruction}.
\begin{figure}
\newcommand{\wsd}{1.02\columnwidth}  % width solar decomp
\newcommand{\hsd}{4cm}  % height solar decomp
\newcommand{\srd}{figures/decomposition/11-Feb-02-solar-s}  % solar decomp results dir
\newcommand{\mbs}{\hspace{-0.3cm}}  % move back
\begin{tabular}{c}
\mbs \includegraphics[width=\wsd,height=\hsd]{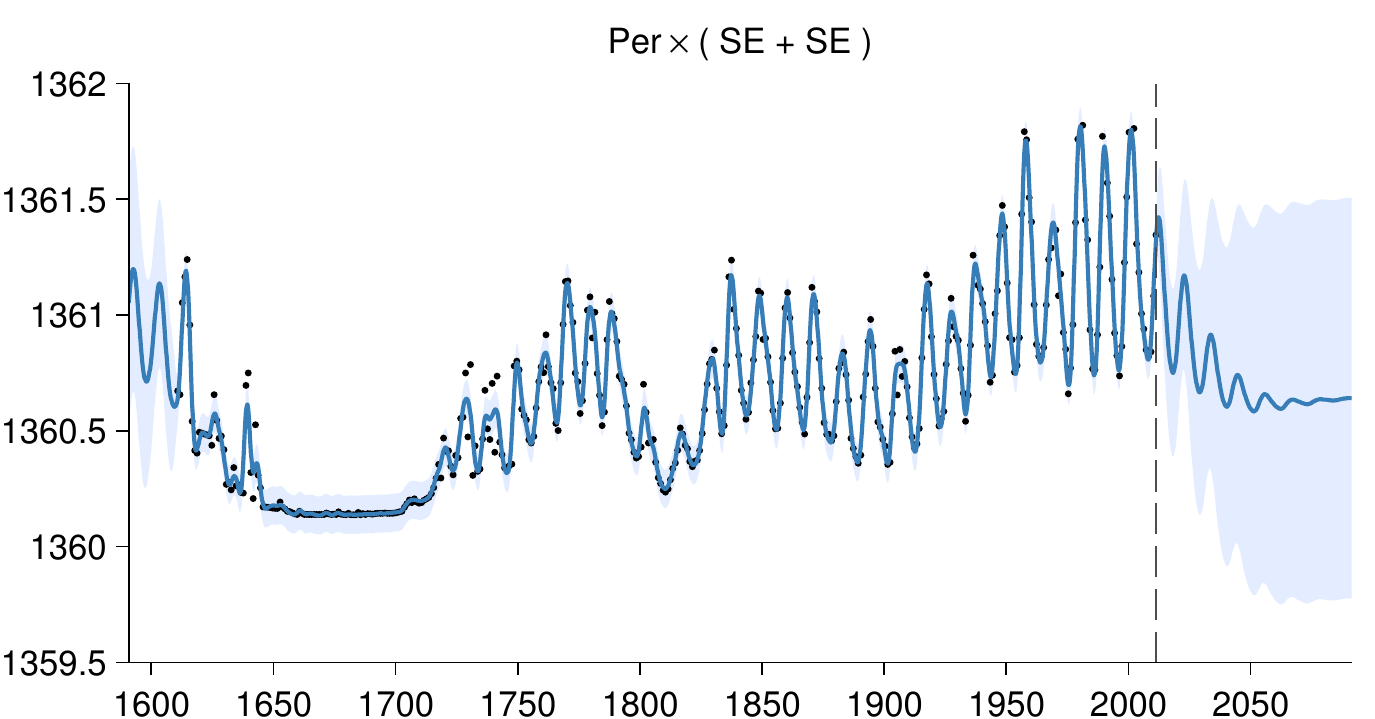} \\
\mbs \includegraphics[width=\wsd,height=\hsd]{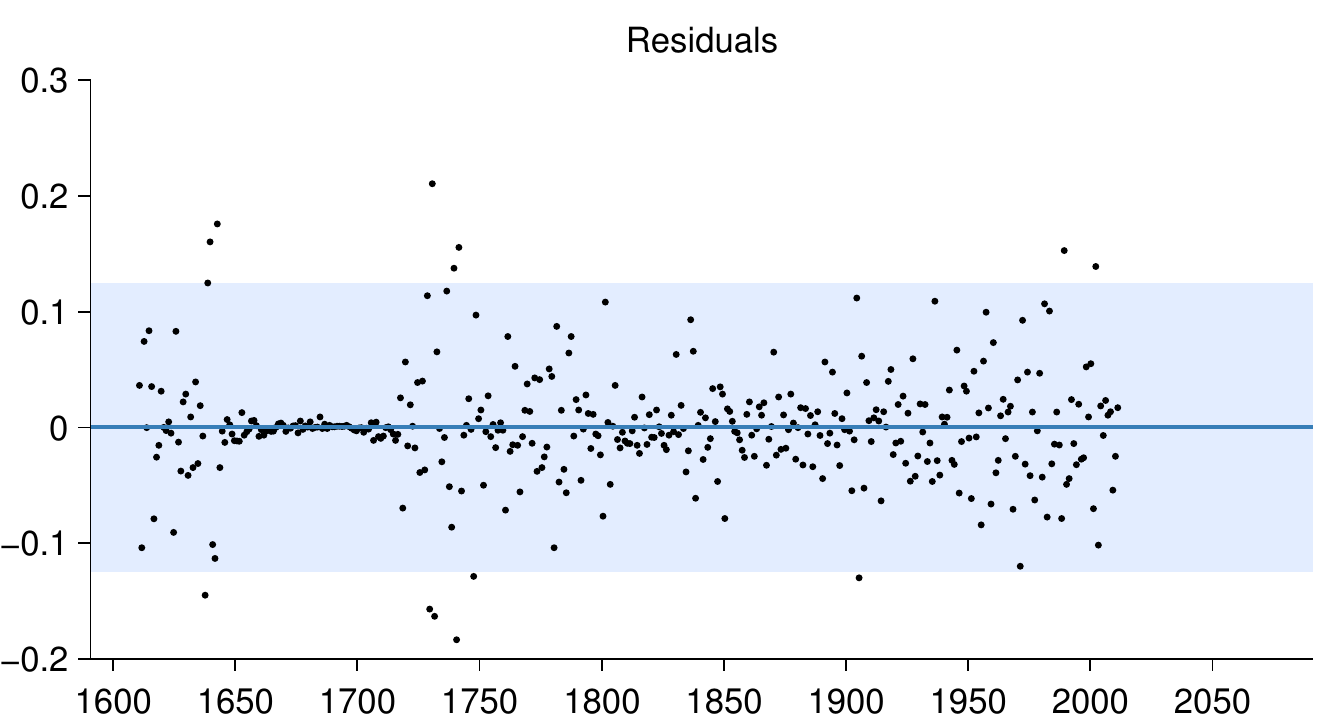}
\end{tabular}
\caption{Full posterior and residuals on the solar irradiance dataset.}
\label{fig:solar_decomp}
\end{figure}

The posterior and residuals of the learned kernel are shown in figure \ref{fig:solar_decomp}.
%The composite kernel captures the periodic structure in the data, but does not capture the flat structure from 1645 to 1715 during which sunspots were extremely rare.
%but it misses out on another aspect of the data: the flat period from 1645 to 1715 which contains no periodicity and has much smaller variance than the rest of the dataset.
%This corresponds to the Maunder minimum, a period in which sunspots were extremely rare.

None of the models in our search space are capable of parsimoniously representing the lack of variation from 1645 to 1715. %, since all of the base kernels apart from \kLin{} are stationary.
%, and it is hard to see how Lin would help in modeling this structure.
%
%
Despite this, our approach fails gracefully: the learned kernel still captures the periodic structure, and the quickly growing posterior variance demonstrates that the model is uncertain about long term structure.

%It is likely that our framework could be extended to capture nonstationary structure by adding a changepoint kernel \TBD{cite?}.

%\paragraph{Multidimensional decomposition}  Decomposition of multi-dimensional posteriors into sums of one-dimensional functions is possible as well, and was demonstrated in \cite{duvenaud2011additive11}.  \fTBD{Probably citing myself too much...}

\section{Validation on synthetic data}
\label{sec:synthetic}

\begin{table*}[ht!]
\caption{{\small
%Kernels used to generate synthetic data, dimensionality, $D$, of the input space, and the estimated kernels.% at different signal to noise ratios (SNR).
Kernels chosen by our method on synthetic data generated using known kernel structures. $D$ denotes the dimension of the functions being modeled.  SNR indicates the signal-to-noise ratio. Dashes - indicate no structure.
%The two kernels marked with an asterisk * indicate when the search procedure inferred extraneous structure.
}}
\label{tbl:synthetic}
\begin{center}
{\small
\begin{tabular}{c c | c c c}
True Kernel & $D$ & SNR = 10 & SNR = 1 & \hspace{-1cm} SNR = 0.1 \\
\hline
$\SE + \RQ$                               & 1 
                                              & $\SE$
                                              & $\SE \times \Per$
                                              & $\SE$
                                              \\
$\Lin \times \Per$                        & 1 
                                              & $\Lin \times \Per$
                                              & $\Lin \times \Per$
                                              & $\SE$
                                              \\
$\SE_1 + \RQ_2$                           & 2 
                                              & $\SE_1 + \SE_2$
                                              & $\Lin_1 + \SE_2$ 
                                              & $\Lin_1$
                                              \\
$\SE_1 + \SE_2 \times \Per_1 + \SE_3$     & 3 
                                              & $\SE_1 + \SE_2 \times \Per_1 + \SE_3$
                                              & $\SE_2 \times \Per_1 + \SE_3$
                                              & -
                                              \\
$\SE_1 \times \SE_2$                      & 4 
                                              & $\SE_1 \times \SE_2$
                                              & $\Lin_1 \times \SE_2$
                                              & $\Lin_2$
                                              \\
$\SE_1 \times \SE_2 + \SE_2 \times \SE_3$ & 4 
                                              & $\SE_1 \times \SE_2 + \SE_2 \times \SE_3$
                                              & $\SE_1 + \SE_2 \times \SE_3$
                                              & $\SE_1$
                                              \\
$(\SE_1 + \SE_2) \times (\SE_3 + \SE_4)$  & 4 
                                              & $(\SE_1 + \SE_2) \times (\SE_3\times\Lin_3\times\Lin_1 + \SE_4)$
                                              & $(\SE_1 + \SE_2) \times \SE_3 \times \SE_4$
                                              & -
                                              \\
\end{tabular}
}
\end{center}
\end{table*}

We validated our method's ability to recover known structure on a set of synthetic datasets.
For several composite kernel expressions, we constructed synthetic data by first sampling 300 points uniformly at random, then sampling function values at those points from a \gp{} prior.
%We then added \iid Gaussian noise to the function values, with variance chosen such that the standard deviation of the noise $\sigma_n$ relative to the sample variance of the function was 0.1.
We then added \iid Gaussian noise to the functions, at various signal-to-noise ratios (SNR).

Table~\ref{tbl:synthetic} lists the true kernels we used to generate the data.  Subscripts indicate which dimension each kernel was applied to.  Subsequent columns show the dimensionality $D$ of the input space, and the kernels chosen by our search for different SNRs.
Dashes - indicate that no kernel had a higher marginal likelihood than modeling the data as \iid Gaussian noise. % (\ie equivalent to a constant kernel).

%\NA{
For the highest SNR, the method finds all relevant structure in all but one test.
The reported additional linear structure is explainable by the fact that functions sampled from \kSE{} kernels with long length scales occasionally have near-linear trends.
%}
%
%\NA{
%As the SNR lowers, the kernel expressions generally become simpler.
As the noise increases, our method generally backs off to simpler structures.
%, rather than over-fitting.
%Two kernels have been marked with an asterisk indicating that the search has inferred more complex and erroneous structure.
%For the first data set with $\textrm{SNR} = 1$, the search erroneously infers periodic structure.
%On examining the data set, and the posterior fit of this kernel, the data could certainly be argued to have ambiguous structure.
%}

% --- Automatically generated by resultsToLatex2.m ---
% Exported at 28-Jan-2013 15:53:45
\begin{table*}[ht!]
\vspace{-0.2cm}
\caption{{\small
Comparison of multidimensional regression performance. Bold results are not significantly different from the best-performing method in each experiment, in a paired t-test with a $p$-value of 5\%.
}}
\label{tbl:Regression Mean Squared Error}
{\small
\begin{center}
\begin{tabularx}{\textwidth}{l | XXXXX | XXXXX}
 & \multicolumn{5}{c}{Mean Squared Error (MSE)} & \multicolumn{5}{c}{Negative Log-Likelihood} \\
 Method & bach  & concrete  & puma  &  servo & housing
& bach  & concrete  & puma  &  servo & housing
\\ \hline
Linear Regression 
& $1.031$ & $0.404$ & $0.641$ & $0.523$ & $0.289$ 
& $2.430$ & $1.403$ & $1.881$ & $1.678$ & $1.052$ \\
GAM 
& $1.259$ & $0.149$ & $0.598$ & $0.281$ & $0.161$ 
& $1.708$ & $0.467$ & $1.195$ & $0.800$ & $0.457$ \\
HKL 
& $\mathbf{0.199}$ & $0.147$ & $0.346$ & $0.199$ & $0.151$ 
& - & - & - & - & -\\
\gp{} \kSE{}-ARD 
& $\mathbf{0.045}$ & $0.157$ & $0.317$ & $0.126$ & $\mathbf{0.092}$ 
& $\mathbf{-0.131}$ & $0.398$ & $0.843$ & $0.429$ & $0.207$ \\
\gp{} Additive 
& $\mathbf{0.045}$ & $\mathbf{0.089}$ & $\mathbf{0.316}$ & $\mathbf{0.110}$ & $0.102$ 
& $\mathbf{-0.131}$ & $\mathbf{0.114}$ & $\mathbf{0.841}$ & $\mathbf{0.309}$ & $0.194$ \\
\hline
Structure Search 
& $\mathbf{0.044}$ & $\mathbf{0.087}$ & $\mathbf{0.315}$ & $\mathbf{0.102}$ & $\mathbf{0.082}$ 
& $\mathbf{-0.141}$ & $\mathbf{0.065}$ & $\mathbf{0.840}$ & $\mathbf{0.265}$ & $\mathbf{0.059}$
\end{tabularx}
\end{center}
}
\end{table*}
% End automatically generated LaTeX

%\input{tables/regression_results_ext_combined.tex}
%\input{tables/regression_results_ext_mse.tex}
%\input{tables/regression_results_ext_nll.tex}

\section{Quantitative evaluation}
\label{sec:quantitative}

%In addition to producing highly interpretable decompositions of regression functions, our method also produces state of the art predictions in both extrapolation and interpolation tasks.
In addition to the qualitative evaluation in section \ref{sec:time_series}, we investigated quantitatively how our method performs on both extrapolation and interpolation tasks.

\subsection{Extrapolation}

\begin{figure}
\centering
\begin{tabular}{c}
\hspace{-0.5cm} \includegraphics[width=0.95\columnwidth,height=7cm]{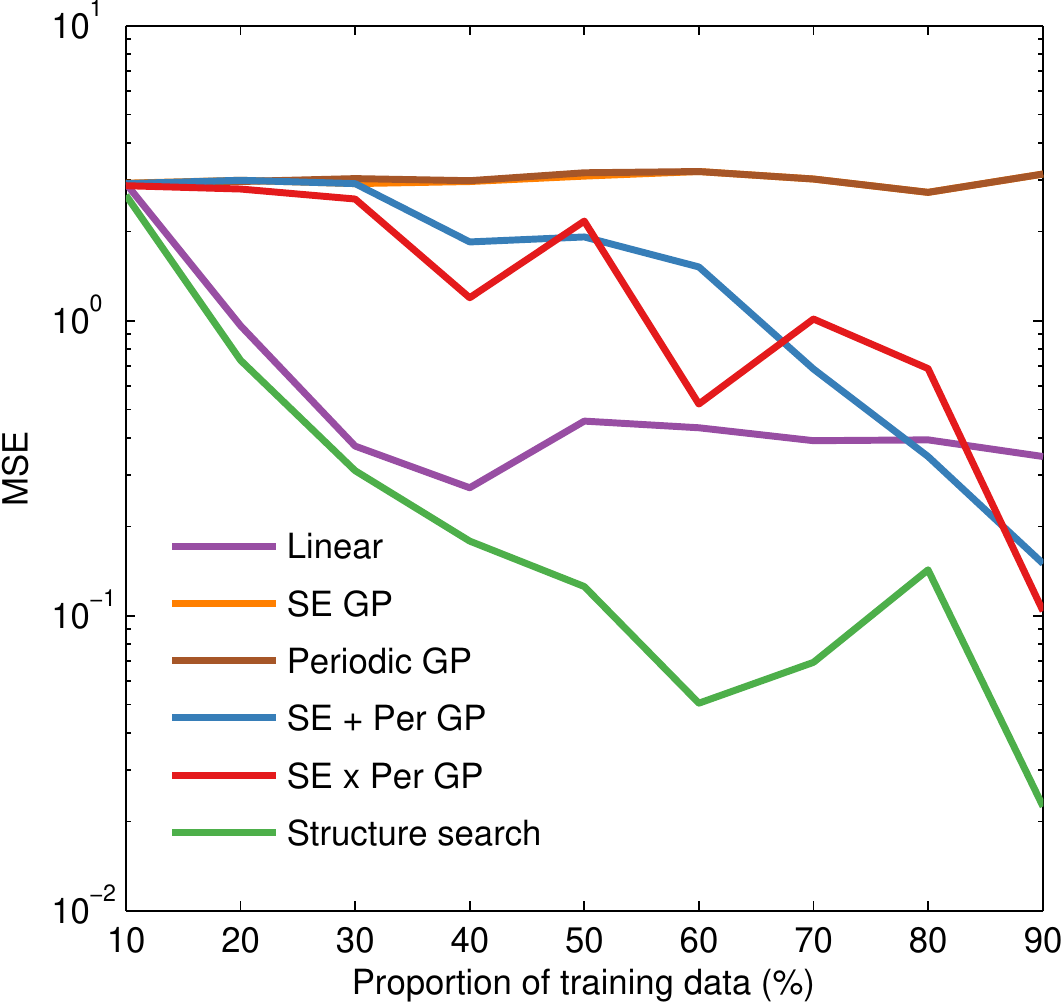}
\end{tabular}
\caption{Extrapolation performance on the airline dataset.  We plot test-set MSE as a function of the fraction of the dataset used for training. 
%\TBD{RBG: Are we using a fixed test set, or the complement of the training set?  It seems like we should do the former, so that the results are less noisy.}
}
\label{fig:extrapolation}
\end{figure}

We compared the extrapolation capabilities of our model against standard baselines\footnotemark.
Dividing the airline dataset into contiguous training and test sets, we computed the predictive mean-squared-error (MSE) of each method.
We varied the size of the training set from the first 10\% to the first 90\% of the data.

Figure \ref{fig:extrapolation} shows the learning curves of linear regression, a variety of fixed kernel family \gp{} models, and our method.  
\gp{} models with only \kSE{} and \kPer{} kernels did not capture the long-term trends, since the best parameter values in terms of \gp{} marginal likelihood only capture short term structure. 
Linear regression approximately captured the long-term trend, but quickly plateaued in predictive performance.
The more richly structured \gp{} models (${\kSE + \kPer}$ and ${\kSE \times \kPer}$) eventually captured more structure and performed better, but the full structures discovered by our search outperformed the other approaches in terms of predictive performance for all data amounts.
%In contrast, a \gp{} with an SE or RQ kernel has unbounded capacity.  However, simply having unbounded capacity does not necessarily translate into the ability to extrapolate, as demonstrate in Figure \ref{fig:extrapolation}.  Only by increasing the amount of structure expressible in a model can we capture the regularities in the data that allow long-range extrapolation.

\footnotetext{
%\NA{
In one dimension, the predictive means of all baseline methods in table \ref{tbl:Regression Mean Squared Error} are identical to that of a \gp{} with an $\kSE{}$ kernel.}
%}

%\fTBD{Josh: Can you write about the Blessing of abstraction, and doing lots with a small amount of data?  This might become apparent from plotting the extraplolations.}
%\fTBD{RBG: We could make this point as part of the learning curves for extrapolation by marking the points at which each additional aspect of the structure is found.}

\subsection{High-dimensional prediction}

To evaluate the predictive accuracy of our method in a high-dimensional setting, we extended the comparison of \cite{duvenaud2011additive11} to include our method.
We performed 10 fold cross validation on 5 datasets
\footnote{The data sets had dimensionalities ranging from 4 to 13, and the number of data points ranged from 150 to 450.} comparing 5 methods in terms of MSE and predictive likelihood.
%\NA{
Our structure search was run up to depth 10, using the \SE{} and \RQ{} base kernel families.
%different subsets of the base kernel families.
%}

The comparison included three methods with fixed kernel families: Additive \gp{}s, Generalized Additive Models (GAM), and a \gp{} with a standard \kSE{} kernel using Automatic Relevance Determination (\gp{} \kSE{}-ARD).  Also included was the related kernel-search method of Hierarchical Kernel Learning (HKL).
%We compared 5 methods on 5 datasets in terms of mean-squared (MSE) error and predictive likelihood, averaging across 10 data folds.

Results are presented in table \ref{tbl:Regression Mean Squared Error}.  Our method outperformed the next-best method in each test, although not substantially.

All \gp{} hyperparameter tuning was performed by automated calls to the GPML toolbox\footnote{Available at 
\href{http://www.gaussianprocess.org/gpml/code/}
{\texttt{www.gaussianprocess.org/gpml/code/}}
}; Python code to perform all experiments is available on github\footnote{
%\texttt{github.com/jamesrobertlloyd/gp-structure-search}
\href{http://www.github.com/jamesrobertlloyd/gp-structure-search}
{\texttt{github.com/jamesrobertlloyd/gp-structure-search}}
}.

%$k = 1$, $D = 8$, kernels are SE and RQ (\TBD{currently running other experiments that may be more canonical}).
%We have extended the comparison of \cite{duvenaud2011additive11} to include our method.

%Some points for discussion
%\begin{itemize}
%\item Experiments just using SE kernel can outperform additive kernel surprisingly. This is presumably a regularisation effect of using a finite depth search and/or BIC. We could make this a more Bayesian result (i.e.~more a property of the model) by placing a prior on kernels that depends on the number of components.
%\item Need to discuss design choices e.g.~$k$, depth of search, base kernels.
%\end{itemize}

\section{Discussion}

\begin{quotation}
``It would be very nice to have a formal apparatus that gives us some `optimal' way of recognizing unusual phenomena and inventing new classes of hypotheses that are most likely to contain the true one; but this remains an art for the creative human mind.''
% In trying to practice this art, the Bayesian has the advantage because his formal apparatus already developed gives him a clearer picture of what to expect, and therefore a sharper perception for recognizing the unexpected.

\defcitealias{Jaynes85highlyinformative}{E. T.  Jaynes, 1985}
%\hspace*{\fill}\citet{Jaynes85highlyinformative}
\hspace*{\fill}\citetalias{Jaynes85highlyinformative}

%\emph{ E.T. Jaynes  from the last paragraph (p. 351) of his 1985 paper, “Highly Informative Priors.”}

\end{quotation}

Towards the goal of automating the choice of kernel family, we introduced a space of composite kernels defined compositionally as sums and products of a small number of base kernels.  
The set of models included in this space includes many standard regression models.
We proposed a search procedure for this space of kernels which parallels the process of scientific discovery.

We found that the learned structures are often capable of accurate extrapolation in complex time-series datasets, and are competitive with widely used kernel classes and kernel combination methods on a variety of prediction tasks.
The learned kernels often yield decompositions of a signal into diverse and interpretable components, enabling model-checking by humans.  %provides an additional degree of reassurance that the learned structure reflects the world.
We believe that a data-driven approach to choosing kernel structures automatically can help make nonparametric regression and classification methods accessible to non-experts.

\subsubsection*{Acknowledgements}
We thank Carl Rasmussen and Andrew G. Wilson for helpful discussions.  This work was funded in part by NSERC, EPSRC grant EP/I036575/1, and Google.

\appendix
\section*{Appendix}
\label{appendix}
\paragraph{Kernel definitions}
For scalar-valued inputs, the squared exponential (\kSE), periodic (\kPer), linear (\kLin), and rational quadratic (\kRQ) kernels are defined as follows:
\begin{eqnarray*}
\kernel_{\SE}(\inputVar, \inputVar') =& \sigma^2\exp\left(-\frac{(\inputVar - \inputVar')^2}{2\ell^2}\right) \\
\kernel_{\Per}(\inputVar, \inputVar') =& \sigma^2\exp\left(-\frac{2\sin^2 (\pi(\inputVar - \inputVar')/p)}{\ell^2}\right) \\
\kernel_{\Lin}(\inputVar, \inputVar') =& \sigma_b^2 + \sigma_v^2(\inputVar - \ell)(\inputVar' - \ell) \\
\kernel_{\RQ}(\inputVar, \inputVar') =& \sigma^2\left( 1 + \frac{(\inputVar - \inputVar')^2}{2 \alpha \ell^2} \right)^{-\alpha}
\end{eqnarray*}

\paragraph{Posterior decomposition}
\label{sec:decomposing}
We can analytically decompose a \gp{} posterior distribution over additive components using the following identity:
%Given a composite kernel consisting of a sum, % of several simpler kernels,
%we can analytically infer a decomposition of a function into a superposition of component functions.
%Formally,\fTBD{Old decomp commented out - but happy to have reintroduced}
%For simplicity, consider the case of a sum of two kernels.
%let ${\function = \sum\function_n}$, where ${\function_n \sim \GP( 0, k_n)}$.
The conditional distribution of a Gaussian vector $\vf_1$ conditioned on its sum with another Gaussian vector $\vf = \vf_1 + \vf_2$ where $\vf_1 \sim \Nt{\vmu_1}{\vK_1}$ and $\vf_2 \sim \Nt{\vmu_2}{\vK_2}$ is given by
%The conditional distribution of a vector of component values $\vf_n$ conditioned on observations $\vf$ is given by
%\[
%\vf_1 | \vf \sim \mathcal{N} \Big( & \boldsymbol\mu_1 + \vk_1\tra (\vK_1 + \vK_2)\inv \left( \vf - \boldsymbol\mu_1 - \boldsymbol\mu_2 \right), \nonumber \\
%& \vK_1 - \vk_1\tra (\vK_1 + \vK_2)\inv \vk_1 \Big).
%\vf_n \given \vf \sim \mathcal{N} \Big( \vK_n (\vK_n + \vK_{-n}\inv)\inv \vf, (\vK_n\inv + \vK_{-n}\inv)\inv \Big).
%\]
%where ${\vK_{n,ij} = k_n(x_i,x_j),\,\,\vK_{-n,ij} = \sum_{m\neq n}k_m(x_i,x_j)}$. 
%
\begin{align}
\vf_1 | \vf \sim \mathcal{N} \big( & \vmu_1 + \vK_1\tra (\vK_1 + \vK_2)\inv \left( \vf - \vmu_1 - \vmu_2 \right), \nonumber \\
& \vK_1 - \vK_1\tra (\vK_1 + \vK_2)\inv \vK_1 \big) \nonumber .
\end{align}
%
%and the covariance between the two components, conditioned on their sum is given by:
%\begin{align}
%\cov(\vf_1^\star, \vf_2^\star) | \vf = \vk_1^{\star\tra} (\vK_1 + \vK_2)\inv \vk_2^\star
%\end{align}
%Derivations can be found in the supplementary material.
%where ${\vK_{ij} = k(x_i,x_j)}$. 

\newpage
\bibliographystyle{format/icml2013}
\bibliography{gpss}
\end{document}



\onecolumn

\icmltitlerunning{Structure Search Supplementary Material}

\icmltitle{Supplementary Material for
\\
Kernel Structure Discovery in Gaussian Process Models}

%\icmlauthor{Anonymous Author 1}
%\icmladdress{Anonymous Institution}

%\subsection{Real datasets}

\appendix

\section{Derivation of Component Marginal Variance}

Let us assume that our function $\vf$ is a sum of two functions, $\vf_1$ and $\vf_2$, where $\vf = \vf_1 + \vf_2$.  If $\vf_1$ and $\vf_2$ are a priori independent, and $\vf_1 \sim \gp( \vmu_1, k_1)$ and $\vf_2 \sim \gp( \vmu_2, k_2)$, then
\begin{align}
\left[ \begin{array}{c} \vf_1 \\ \vf_1^\star \\ \vf_2 \\ \vf_2^\star \\ \vf \\ \vf^\star \end{array} \right]
\sim
\Nt{\left[ \begin{array}{c} \vmu_1 \\ \vmu_1^\star \\ \vmu_2 \\ \vmu_2^\star \\ \vmu_1 + \vmu_2 \\ \vmu_1^\star + \vmu_2^\star \end{array} \right]
}
{\left[ \begin{array}{cccccc} 
\vk_1 & \vk_1^\star & 0 & 0 & \vk_1 & \vk_1^\star \\ 
\vk_1^\star & \vk_1^{\star\star} & 0 & 0 & \vk_1^\star & \vk_1^{\star\star} \\
0 & 0 & \vk_2 & \vk_2^\star & \vk_2 & \vk_2^\star \\ 
0 & 0 & \vk_2^\star & \vk_2^{\star\star} & \vk_2^\star & \vk_2^{\star\star} \\
\vk_1 & \vk_1^\star & \vk_2 & \vk_2^\star & \vk_1 + \vk_2 & \vk_1^\star + \vk_2^\star \\ 
\vk_1^\star & \vk_1^{\star\star}  & \vk_2^\star & \vk_2^{\star\star}  & \vk_1^\star + \vk_2^\star & \vk_1^{\star\star} + \vk_2^{\star\star}\\
\end{array} \right]
}
\end{align}
where $\vk_1 = k_1( \vX, \vX )$ and $\vk_1^\star = k_1( \vX^\star, \vX )$. 

By the formula for Gaussian conditionals:
\begin{align}
\vx_A | \vx B \sim \Nt{\vmu_A + \vSigma_{AB} \vSigma_{BB}\inv \left( \vx_B - \vmu_B \right) }
{\vSigma_{AA} - \vSigma_{AB} \vSigma_{BB}\inv \vSigma_{BA} },
\end{align}
we get that the conditional variance of a Gaussian conditioned on its sum with another Gaussian is given by
\begin{align}
\vf_1^\star | \vf \sim \Nt{\vmu_1^\star + \vk_1^{\star\tra} (\vK_1 + \vK_2)\inv \left( \vf - \vmu_1 - \vmu_2 \right) }
{\vk_1^{\star\star} - \vk_1^{\star\tra} (\vK_1 + \vK_2)\inv \vk_1^\star }.
\end{align}

The covariance between the two components, conditioned on their sum is given by:
\begin{align}
\cov(\vf_1^\star, \vf_2^\star) | \vf = - \vk_1^{\star\tra} (\vK_1 + \vK_2)\inv \vk_2^\star
\end{align}

These formulae express the posterior model uncertainty about different components of the signal, integrating over the possible configurations of the other components.

\section{Details of Search Algorithm}

Formally\fTBD{Maybe Roger is well placed to write this?}, we start with a collection of base kernels applied to each input dimension individually; $\kernel_i$ denotes a kernel applied to input dimension $i$.
We denote sums and products of kernels as $\SumKernel(\expression,\ldots)$ and $\ProductKernel(\expression,\ldots)$ where $\expression$ represents an arbitrary kernel expression.
We then generate new expressions by repeatedly applying the following production rules to base kernels, sum kernels and product kernel within an expression.
\begin{center}
\begin{tabular}{rccc}
\textrm{Replacement} & $\kernel_i$ & $\to$ & $\kernel'_i$\\% & $\forall\, \kernel' $\\
\textrm{Addition} & $\kernel_i$ & $\to$ & $\kernel_i + \kernel'_j$\\% & $\forall\, j,\kernel' $\\
& $\SumKernel(e,\ldots)$ & $\to$ & $\SumKernel(e,\ldots,\kernel'_j)$\\% & $\forall\, j,\kernel' $\\
& $\ProductKernel(e,\ldots)$ & $\to$ & $\SumKernel(\ProductKernel(e,\ldots),\kernel'_j)$\\% & $\forall\, j,\kernel' $\\
\textrm{Multiplication} & $\kernel_i$ &  $\to$ & $\kernel_i \times \kernel'_j$\\% & $\forall\, j,\kernel'$\\
& $\SumKernel(e,\ldots)$ & $\to$ & $\ProductKernel(\SumKernel(e,\ldots),\kernel'_j)$\\% & $\forall\, j,\kernel' $\\
& $\ProductKernel(e,\ldots)$ & $\to$ & $\ProductKernel(e,\ldots,\kernel'_j)$\\% & $\forall\, j,\kernel' $\\
\end{tabular}
\end{center}
For example, applying the multiplication production rule to the sum in the expression $\SumKernel(\kernel_\textrm{SE},\kernel_\textrm{SE})$ could result in the new expression $\ProductKernel(\SumKernel(\kernel_\textrm{SE},\kernel_\textrm{SE}),\kernel_\textrm{PE})$ \ie a kernel representing a smooth function with two characteristic scales of variation is transformed into a kernel representing a locally periodic function with two scales of variation.

\paragraph{Remark on notation} The $\SumKernel,\ProductKernel$ notation is useful algorithmically and to describe the production rules but elsewhere we will use more conventional algebraic notation \ie the two kernels above would be written as $\kernel_\textrm{SE} + \kernel_\textrm{SE}$ and ${(\kernel_\textrm{SE} + \kernel_\textrm{SE}) \times \kernel_\textrm{PE}}$.

%\section{Decompositions}

%\begin{figure}
%\includegraphics[width=10cm]{../figures/decomposition/01-airline/01-airline_all.pdf}
%\caption{The complete posterior on the Airline dataset.}
%\label{fig:mauna_all}
%\end{figure}

%\newcommand{\fw}{8cm}
%\begin{figure*}
%\centering
%\begin{tabular}{cc}
% \includegraphics[width=\fw]{../figures/decomposition/01-airline/01-airline_7.pdf} &  \includegraphics[width=\fw]{../figures/decomposition/01-airline/01-airline_8.pdf} \\
%  \includegraphics[width=\fw]{../figures/decomposition/01-airline/01-airline_5.pdf} &  \includegraphics[width=\fw]{../figures/decomposition/01-airline/01-airline_6.pdf} \\
%   \includegraphics[width=\fw]{../figures/decomposition/01-airline/01-airline_3.pdf} &  \includegraphics[width=\fw]{../figures/decomposition/01-airline/01-airline_4.pdf} \\
%    \includegraphics[width=\fw]{../figures/decomposition/01-airline/01-airline_1.pdf} &  \includegraphics[width=\fw]{../figures/decomposition/01-airline/01-airline_2.pdf}
%\end{tabular}
%\caption{Automatic decomposition of airline data.
%}
%\label{fig:kernels}
%\end{figure*}

\section{Complete listings of chosen kernels}

Below, we show the best kernel structure found for each fold of the datasets used in the results tables.

%\input{"tables/kernels.tex"}
%\input{"tables/kernels2.tex"}
%\input{"tables/kernels3.tex"}
%% --- Automatically generated by latex.py ---
% Exported at 2013-01-28 10:34:20.566777
\begin{table*}[h!]
\begin{center}
\begin{tabular}{l | l l l}
 Dataset  & \rotatebox{0}{ NLL }  & \rotatebox{0}{ Kernel }  \\ \hline
bachsynthr200fold10of10result & $ 24.9 $ & $ SE_{0} \times SE_{1} \times SE_{2} \times SE_{3} $ \\
bachsynthr200fold1of10result & $ 19.0 $ & $ SE_{0} \times SE_{1} \times SE_{2} \times SE_{3} $ \\
bachsynthr200fold2of10result & $ 21.0 $ & $ SE_{0} \times SE_{1} \times SE_{2} \times SE_{3} $ \\
bachsynthr200fold3of10result & $ 21.1 $ & $ SE_{0} \times SE_{1} \times SE_{2} \times SE_{3} $ \\
bachsynthr200fold4of10result & $ 18.2 $ & $ SE_{0} \times SE_{1} \times SE_{2} \times SE_{3} $ \\
bachsynthr200fold5of10result & $ 22.4 $ & $ SE_{0} \times SE_{1} \times SE_{2} \times SE_{3} $ \\
bachsynthr200fold6of10result & $ 20.2 $ & $ SE_{0} \times SE_{1} \times SE_{2} \times SE_{3} $ \\
bachsynthr200fold7of10result & $ 21.0 $ & $ SE_{0} \times SE_{1} \times SE_{2} \times SE_{3} $ \\
bachsynthr200fold8of10result & $ 17.9 $ & $ SE_{0} \times SE_{1} \times SE_{2} \times SE_{3} $ \\
bachsynthr200fold9of10result & $ 22.2 $ & $ SE_{0} \times SE_{1} \times SE_{2} \times SE_{3} $ \\
rconcrete500fold10of10result & $ 103.8 $ & $ RQ_{1} \times RQ_{4} \times SE_{6} \times RQ_{7} \times \left( SE_{7} + RQ_{5} \times \left( RQ_{0} + SE_{3} \right) \right) $ \\
rconcrete500fold1of10result & $ 76.2 $ & $ RQ_{1} \times RQ_{4} \times SE_{6} \times RQ_{7} \times \left( RQ_{0} \times RQ_{3} + SE_{6} \times SE_{7} \right) $ \\
rconcrete500fold2of10result & $ 114.8 $ & $ SE_{1} \times RQ_{4} \times RQ_{6} \times \left( SE_{0} \times RQ_{3} + RQ_{5} \times SE_{7} \right) $ \\
rconcrete500fold3of10result & $ 111.7 $ & $ RQ_{1} \times SE_{6} \times \left( RQ_{4} + RQ_{6} \right) \times \left( RQ_{7} + RQ_{0} \times \left( RQ_{3} + SE_{7} \right) \right) $ \\
rconcrete500fold4of10result & $ 123.9 $ & $ SE_{1} \times RQ_{4} \times SE_{6} \times RQ_{7} \times \left( SE_{7} + SE_{0} \times RQ_{3} \right) $ \\
rconcrete500fold5of10result & $ 100.0 $ & $ SE_{1} \times RQ_{4} \times RQ_{5} \times SE_{6} \times RQ_{7} \times \left( SE_{7} + RQ_{0} \times RQ_{3} \right) $ \\
rconcrete500fold6of10result & $ 114.9 $ & $ SE_{0} \times SE_{1} \times RQ_{4} \times RQ_{6} \times \left( SE_{7} + SE_{3} \times SE_{5} \right) $ \\
rconcrete500fold7of10result & $ 112.9 $ & $ SE_{1} \times RQ_{4} \times RQ_{6} \times RQ_{7} \times \left( SE_{7} + RQ_{0} \times RQ_{3} \right) $ \\
rconcrete500fold8of10result & $ 89.4 $ & $ SE_{0} \times RQ_{1} \times RQ_{3} \times RQ_{4} \times RQ_{6} \times \left( RQ_{7} + RQ_{0} \times SE_{5} \right) $ \\
rconcrete500fold9of10result & $ 100.7 $ & $ SE_{0} \times SE_{1} \times RQ_{4} \times RQ_{6} \times \left( RQ_{7} + RQ_{0} \times RQ_{3} \right) $ \\
rhousingfold10of10result & $ 103.0 $ & $ RQ_{4} \times SE_{5} \times SE_{6} \times SE_{9} \times RQ_{11} \times SE_{12} $ \\
rhousingfold1of10result & $ 106.2 $ & $ SE_{0} \times RQ_{4} \times SE_{5} \times SE_{6} \times SE_{9} \times SE_{11} \times SE_{12} $ \\
rhousingfold2of10result & $ 106.8 $ & $ RQ_{4} \times SE_{5} \times SE_{6} \times SE_{9} \times SE_{12} \times \left( SE_{0} + SE_{11} \right) $ \\
rhousingfold3of10result & $ 117.6 $ & $ RQ_{4} \times SE_{5} \times SE_{6} \times SE_{9} \times SE_{12} \times \left( SE_{0} + SE_{11} \right) $ \\
rhousingfold4of10result & $ 103.5 $ & $ SE_{0} \times RQ_{4} \times SE_{5} \times SE_{6} \times SE_{11} \times SE_{12} \times \left( SE_{4} + SE_{9} \right) $ \\
rhousingfold5of10result & $ 96.8 $ & $ SE_{0} \times RQ_{4} \times SE_{5} \times SE_{6} \times SE_{9} \times SE_{11} \times SE_{12} $ \\
rhousingfold6of10result & $ 102.6 $ & $ SE_{0} \times RQ_{4} \times SE_{5} \times SE_{6} \times SE_{9} \times RQ_{11} \times SE_{12} $ \\
rhousingfold7of10result & $ 113.7 $ & $ SE_{0} \times SE_{5} \times SE_{9} \times SE_{11} \times SE_{12} \times \left( RQ_{4} + SE_{6} \right) $ \\
rhousingfold8of10result & $ 96.8 $ & $ RQ_{4} \times SE_{5} \times SE_{6} \times SE_{9} \times SE_{11} \times SE_{12} $ \\
rhousingfold9of10result & $ 88.1 $ & $ RQ_{4} \times SE_{5} \times SE_{6} \times SE_{9} \times SE_{12} \times \left( SE_{0} + SE_{11} \right) $ \\
rpumadyn512fold10of10result & $ 397.8 $ & $ SE_{1} \times SE_{2} $ \\
rpumadyn512fold1of10result & $ 390.8 $ & $ SE_{1} \times SE_{2} $ \\
rpumadyn512fold2of10result & $ 392.9 $ & $ SE_{1} \times SE_{2} $ \\
rpumadyn512fold3of10result & $ 405.2 $ & $ SE_{1} \times SE_{2} $ \\
rpumadyn512fold4of10result & $ 387.2 $ & $ SE_{1} \times SE_{2} $ \\
rpumadyn512fold5of10result & $ 394.4 $ & $ SE_{1} \times SE_{2} $ \\
rpumadyn512fold6of10result & $ 402.6 $ & $ SE_{1} \times SE_{2} $ \\
rpumadyn512fold7of10result & $ 397.4 $ & $ SE_{1} \times SE_{2} $ \\
rpumadyn512fold8of10result & $ 399.1 $ & $ SE_{1} \times SE_{2} $ \\
rpumadyn512fold9of10result & $ 393.5 $ & $ SE_{2} \times \left( SE_{1} + SE_{3} \right) $ \\
rservofold10of10result & $ 76.2 $ & $ SE_{2} \times \left( SE_{0} + SE_{3} \right) \times \left( SE_{1} + SE_{3} \right) $ \\
rservofold1of10result & $ 78.8 $ & $ RQ_{1} \times SE_{2} \times \left( SE_{0} + SE_{3} \right) $ \\
rservofold2of10result & $ 72.4 $ & $ SE_{2} \times \left( SE_{0} + SE_{3} \right) \times \left( SE_{1} + SE_{3} \right) $ \\
rservofold3of10result & $ 60.8 $ & $ RQ_{1} \times SE_{2} \times \left( SE_{0} + SE_{3} \right) $ \\
rservofold4of10result & $ 77.4 $ & $ RQ_{1} \times SE_{2} \times \left( SE_{0} + SE_{3} \right) $ \\
rservofold5of10result & $ 74.0 $ & $ SE_{2} \times \left( SE_{0} + SE_{3} \right) \times \left( SE_{1} + SE_{3} \right) $ \\
rservofold6of10result & $ 75.4 $ & $ SE_{2} \times \left( SE_{0} + SE_{3} \right) \times \left( SE_{1} + SE_{3} \right) $ \\
rservofold7of10result & $ 62.7 $ & $ SE_{1} \times SE_{2} \times \left( SE_{0} + SE_{3} \right) $ \\
rservofold8of10result & $ 65.0 $ & $ SE_{1} \times SE_{2} \times \left( SE_{0} + SE_{3} \right) $ \\
rservofold9of10result & $ 72.5 $ & $ SE_{0} \times SE_{2} \times SE_{3} \times \left( SE_{0} + SE_{1} \right) $ \\
\end{tabular}
\end{center}
\end{table*}

%% --- Automatically generated by latex.py ---
% Exported at 2013-01-29 09:18:46.732854
\begin{table}[h!]
\begin{center}
\begin{tabular}{l | l l l}
 Dataset  & \rotatebox{0}{ NLL }  & \rotatebox{0}{ Kernel }  \\ \hline
bachsynthr200fold10of10result & $  6.0 $ & $ \left( LN_{0} + SE_{1} \right) \times \left( LN_{1} + LN_{2} + P1_{2} \right) \times \left( LN_{3} + P1_{3} \right) $ \\
bachsynthr200fold1of10result & $ -2.6 $ & $ RQ_{0} \times SE_{1} \times SE_{2} \times \left( LN_{0} + LN_{3} + LN_{1} \times LN_{2} \right) $ \\
bachsynthr200fold2of10result & $  6.1 $ & $ SE_{0} \times \left( LN_{0} + LN_{2} + LN_{3} \right) \times \left( SE_{1} + LN_{2} \right) $ \\
bachsynthr200fold3of10result & $ -1.0 $ & $ \left( SE_{0} + LN_{1} + LN_{2} + LN_{3} \right) \times \left( LN_{0} + LN_{2} + LN_{3} \right) $ \\
bachsynthr200fold4of10result & $ -10.4 $ & $ SE_{3} \times \left( LN_{0} + LN_{1} + LN_{3} \right) \times \left( LN_{0} + P1_{1} + LN_{2} \right) $ \\
bachsynthr200fold5of10result & $  6.6 $ & $ SE_{0} \times \left( LN_{0} + LN_{2} + LN_{3} \right) \times \left( SE_{1} + LN_{3} \right) $ \\
bachsynthr200fold6of10result & $ -1.1 $ & $ SE_{0} \times P1_{1} \times SE_{2} \times \left( LN_{0} + LN_{3} + LN_{1} \times LN_{2} \right) $ \\
bachsynthr200fold7of10result & $  3.8 $ & $ SE_{1} \times \left( SE_{0} + LN_{3} \right) \times \left( LN_{0} + LN_{2} + LN_{3} \right) $ \\
bachsynthr200fold8of10result & $ -3.9 $ & $ P1_{0} \times P1_{1} \times SE_{2} \times \left( LN_{0} + LN_{3} + LN_{1} \times LN_{2} \right) $ \\
bachsynthr200fold9of10result & $  5.6 $ & $ SE_{1} \times \left( SE_{0} + LN_{3} \right) \times \left( LN_{0} + LN_{2} + LN_{3} \right) $ \\
rconcrete500fold10of10result & $ 103.5 $ & $ RQ_{0} \times MT_{3} \times MT_{6} \times \left( LN_{1} + MT_{4} \right) \times \left( MT_{3} + MT_{7} \right) $ \\
rconcrete500fold1of10result & $ 93.3 $ & $ MT_{1} \times RQ_{4} \times MT_{6} \times \left( SE_{0} + P1_{6} \right) \times \left( MT_{7} + MT_{0} \times RQ_{3} \right) $ \\
rconcrete500fold2of10result & $ 114.9 $ & $ MT_{4} \times MT_{6} \times \left( MT_{1} + MT_{4} \right) \times \left( RQ_{0} \times RQ_{3} + RQ_{5} \times MT_{7} \right) $ \\
\end{tabular}
\end{center}
\end{table}

\bibliographystyle{format/icml2013}
\bibliography{gpss}